\documentclass[preprint,review,12pt]{elsarticle}
\usepackage{amssymb}
\usepackage{graphicx}
\usepackage{float}
\usepackage{subfig}
\usepackage{epstopdf}
\usepackage{amsmath,amssymb,amsfonts}
\usepackage{algorithm}
\usepackage{algpseudocode}
\usepackage{textcomp}
\usepackage{xcolor}
\usepackage{array}
\usepackage{balance}
\usepackage{indentfirst}
\usepackage{multirow}
\usepackage{booktabs}
\usepackage{lscape}
\usepackage{wrapfig}
\usepackage{threeparttable}
\usepackage{mathrsfs}

\newcounter{globalprop}
\newtheorem{globalProp}[globalprop]{Proposition} %create gloabal theorem environment

\newcounter{globaldef}
\newtheorem{globalDef}[globaldef]{Definition} %create gloabal definition environment

\newcounter{globalremark}
\newtheorem{globalRem}[globalremark]{Remark} %create gloabal definition environment

\journal{arXiv}

\begin{document}
	
\begin{frontmatter}        
    % \title{Title\tnoteref{label1}}
    \title{Generalized Cauchy-Schwarz Divergence and Its Deep Learning Applications}
    
    \author[1]{Mingfei Lu}
    \author[2]{Chenxu Li}
    \author[3,4]{Shujian Yu}
    \author[4]{Robert Jenssen}
    \author[1]{Badong Chen\corref{cor}}
    \cortext[cor]{Corresponding author. \textit{chenbd@mail.xjtu.edu.cn}(Badong Chen)
                }
    % \cortext[cor2]{\ead{}}
    \address[1]{National Key Laboratory of Human-Machine Hybrid Augmented Intelligence, 
    National Engineering Research Center for Visual Information and Applications, 
    Institute of Artificial Intelligence and Robotics, Xi'an Jiaotong University, Xi'an, 710049, China}
    \address[2]{Xi'an Technological University, Xi'an, 710021, China}
    \address[3]{Vrije Universiteit Amsterdam, HV Amsterdam, 1081, The Netherlands}
    \address[4]{UiT - The Arctic University of Norway, 9037 Troms{\o}, Norway}
        
    \begin{abstract}    
        % % Code is available at~\emph{{https://github.com/LMFLRB/GCSD.git}}. 
        Divergence measures play a central role and become increasingly essential in deep learning, yet efficient measures for multiple (more than two) distributions are rarely explored. This becomes particularly crucial in areas where the simultaneous management of multiple distributions is both inevitable and essential.  Examples include clustering, multi-source domain adaptation or generalization, and multi-view learning, among others. While computing the mean of pairwise distances between any two distributions is a prevalent method to quantify the total divergence among multiple distributions, it is imperative to acknowledge that this approach is not straightforward and necessitates significant computational resources. In this study, we introduce a new divergence measure tailored for multiple distributions named the generalized Cauchy-Schwarz divergence (GCSD). Additionally, we furnish a kernel-based closed-form sample estimator, making it convenient and straightforward to use in various machine-learning applications. Finally, we explore its profound implications in the realm of deep learning by applying it to tackle two thoughtfully chosen machine-learning tasks: deep clustering and multi-source domain adaptation. Our extensive experimental investigations confirm the robustness and effectiveness of GCSD in both scenarios. The findings also underscore the innovative potential of GCSD and its capability to significantly propel machine learning methodologies that necessitate the quantification of multiple distributions.
    \end{abstract}
    
    
\end{frontmatter}
\begin{sloppypar}
%% \linenumbers
%\begin{linenumbers}
%% main text
\section{Introduction}\label{intro}    
The utilization of divergence measures as loss functions to optimize models has gained considerable attention in the field of machine learning~\cite{Basu2011statistical, zhu2022generalized, huang2023highorder}.
It is further popularized by the widespread adoption of deep learning, which has successfully incorporated divergence measures in various domains. 
Examples include deep clustering \cite{xie2016dec, kampffmeyer2019deep, wang2023learning}, domain adaptation or generalization \cite{peng2019moment,hu2020domain, chen2023domain}, generative modeling \cite{kingma2013auto, goodfellow2014generative}, and self-supervised learning \cite{zhuang2022divergence}, among others.     
Over the last decades, substantial efforts have been made in developing a variety of divergence measures for inference and learning~\cite{pardo2018statistical,casella2021statistical}. 
Most existing divergence measures are designed for comparing two distributions, limiting their applicability in scenarios involving multiple ones.    
However, in machine learning, especially the recent deep learning applications, it is often necessary to measure divergence among samples from multiple distributions. 
Computing the mean of pairwise distances between any two distributions is a prevalent method to quantify the total divergence among multiple distributions, it is imperative to acknowledge that this approach is not straightforward and necessitates significant computational resources.

In the field of data clustering, for instance, a common objective is to maximize the total divergence of distributions from the sample or learned features across all clusters~\cite{kampffmeyer2019deep,trosten2021reconsidering}.
Researchers have successfully addressed clustering problems using mean-pairwise divergences, such as Kullback-Leibler divergence (KLD)~\cite{zhou2015learning}, Cauchy-Schwarz divergence (CSD)~\cite{kampffmeyer2019deep}, and Bregman divergence~\cite{banerjee2004information,banerjee2005clustering}.

Another example arises from multi-source domain adaptation (MSDA), where we have access to samples from more than two source domains.
From a representation learning perspective, a prominent idea for MSDA is to align the distributions of the extracted features from all source domains to that of the target domain~\cite{peng2019moment,zhu2019aligning}.
This alignment can be achieved by minimizing the overall divergence between the source and target domains.
Various approaches have been developed to achieve this objective using divergence measures, such as an integrated multi-domain discriminant analysis metric based on average-pairwise maximum mean distance (MMD)~\cite{zhu2019aligning}, and an entropy regularized method utilizing the average-pairwise KLD~\cite{zhao2020domain}. 

However, the average-pairwise divergences do not provide a direct quantification of the total divergence between multiple distributions but involve computing the divergence between each pair of the distributions, leading to potential scalability issues.
Though a few studies have proposed definitions of divergence measures for multiple distributions, including the information radius~\cite{sibson1969information}, barycenter-based dissimilarity \cite{perez1984barycenter}, and $f$-dissimilarity~\cite{gyorfi1978f}, there are few successful applications in the field of machine learning since they were proposed.
Additionally, there is a lack of valid and efficient sample-based estimators for all three categories of generalized divergence measures mentioned above.
This also hinders their application in the field of data-driven machine learning.

In this paper, we propose a novel divergence measure named Generalized Cauchy-Schwarz Divergence (GCSD) for quantifying multiple distributions, taking inspiration from the generalized H\"older’s inequality \cite{finner1992generalization}.
We will show that the GCSD can be elegantly estimated from samples with a closed-form expression, providing a comprehensive and computationally efficient method for comparing multiple distributions.
Notably, the derived estimator is differentiable, facilitating their implementation as loss functions or regularization terms in machine learning tasks.
We finally explore the profound implications of our proposed GCSD in deep learning by applying it to two challenging applications: clustering and domain adaptation. 
The former distinguishes learned features by maximizing divergence amongst them, while the latter aligns data distributions by minimizing the divergence, which are the two main paradigms for applying divergence metrics in machine learning.
Extensive experimental results across diverse benchmark datasets demonstrate remarkable performance, highlighting the effectiveness of the proposed GCSD as well as its broader applicability in other machine learning domains involving multiple data distributions.

Contributions of our work are summarized as follows:
\begin{itemize}
    \item We propose a new generalized divergence measure called the GCSD for comparing multiple distributions. 
    Our theoretical analysis confirms that the GCSD possesses essential properties such as \emph{Non-negativity}, \emph{Symmetry}, and \emph{Projective Invariance}.
    \item We provide a non-parametric estimator for the GCSD, without relying on any distributional assumptions. This approach is both computationally efficient and suitable for use in the field of machine learning.
    \item We empirically demonstrate the effectiveness and robustness of our GCSD estimator through tests on synthetic multi-distribution datasets.
    \item We conduct deep clustering experiments on various image datasets to demonstrate the effectiveness of the GCSD and the divergence-based clustering framework. 
          The success achieved in this task highlights the capability of the GCSD to distinguish features by maximizing generalized divergence measured by GCSD among them.
    \item Our integration of the GCSD into the popular M$^3$SDA (Moment matching for multi-source domain adaptation) framework by Peng et al.~\cite{peng2019moment} performs significantly better on several benchmark datasets than the original M$^3$SDA which relies on pairwise MMD. 
          The success achieved in this task underscores the effectiveness of the GCSD in aligning features by minimizing generalized divergence measured by GCSD among them.
\end{itemize}
% The remainder of this paper is structured as follows: in Section~\ref{sec:retated_work} we introduced the related work

\section{Related Work}
In line with the focus of our work, we briefly review the previous work in the following three areas.
\subsection{Cauchy-Schwarz Divergence}
Motivated by the Cauchy-Schwarz inequality for two square-integrable functions $p({\bf{x}})$ and $q({\bf{x}})$:
\begin{equation}\label{eq:cs_inequality}
    {\left( {\int {p\left( {\bf{x}} \right)q\left( {\bf{x}} \right)d{\bf{x}}} } \right)^2} \le \int {{p^2}\left( {\bf{x}} \right)d{\bf{x}}} \int {{q^2}\left( {\bf{x}} \right)d{\bf{x}}}, 
\end{equation}
CSD~\cite{principe2000learning} defines the distance between two probability density functions by measuring the gap between the left-hand side and right-hand side of Eq.~\eqref{eq:cs_inequality} using the logarithm of their ratio:
\begin{equation}\label{eq:csd}
{D_{{\rm{CS}}}}\left( {\cal {P, Q}} \right) =  - \log \frac{{\int {p\left( {\mathbf{x}} \right)q\left( {\mathbf{x}} \right)d{\mathbf{x}}} }}{{\sqrt {\int {{p^2}\left( {\mathbf{x}} \right)d{\mathbf{x}}} \int {{q^2}\left( {\mathbf{x}} \right)d{\mathbf{x}}} } }},
\end{equation} 
where $p(\mathbf{x})$ and $q(\mathbf{x})$ denote the probability density of distribution $\cal P$ and $\cal Q$ at position $\mathbf{x}$, respectively.

CSD has gained popularity due to its favorable properties, particularly its closed-form expression for mixtures of Gaussians~\cite{kampa2011closed}. 
As a result, it has found successful applications in deep clustering~\cite{kampffmeyer2019deep,trosten2021reconsidering}, disentangled representation learning~\cite{tran2022cauchy}, point-set registration~\cite{sanchez2017group}, among others.
However, the CSD faces limitations when dealing with data from multiple distributions. 
In particular, it exhibits low efficiency with the mean pairwise approach.

Is it possible to develop a multivariate version of the CS divergence that can handle data from multiple distributions while preserving the desirable characteristics of the original version?
We shall delve into these questions and provide answers.

\subsection{Generalized Divergence Measure}
Many applications utilize the mean-pairwise divergence as an alternative measure of generalized divergence. 
Let ${\{{\cal{P}}_t\}}_{t=1}^m$ be a finite set of probability distributions on the measurable space $\cal X$ with the corresponding density functions $\left\{ p_t({\mathbf{x}})\right\}_{t=1}^m$. 
The mean-pairwise divergence is calculated as:
\begin{equation}
    D_{\rm{MP}} = \frac{2}{{m(m - 1)}}{\sum _{i < j}}d({{\cal P}_i},{{\cal P}_j}),
\end{equation}
with $d(\cdot,\cdot)$ an existing divergence measure such as KLD~\cite{zhou2015learning}, CSD~\cite{kampffmeyer2019deep}, MMD~\cite{zhu2019aligning}, and so on.
Although the mean-pairwise divergence is intuitive and easy to understand, it is not a straightforward measure of the general divergence and the traversal operation may give rise to scalability issues.

Three types of generalized divergence measures serve as straightforward methods to quantify the dissimilarity or discrepancy among multiple distributions, including the information radius~\cite{sibson1969information}, barycenter-based dissimilarity \cite{perez1984barycenter}, and $f$-dissimilarity~\cite{gyorfi1978f}.
The information radius is a generalized mean of information divergences between each ${\cal{P}}_t$ and the generalized mean distribution $\bar {\cal P}$~\cite{basseville1996information}, defined as 
\footnote{The term ${{\mathcal P}}_{1:m}$ and the sequence ${\mathcal P}_1, \cdots, {\mathcal P}_m$ are used interchangeably for convenience in this paper.}
\begin{equation}
    D_{\rm{IR}}^{\left( \beta  \right)}\left( {{\mathcal{P}_{1:m}}} \right) = \sum\nolimits_{t=1}^m {{\beta _t}} {d}\left( {{{\cal P}_t},\bar {\cal P}} \right)\;\; s.t.\;\bar {\cal P} = \sum\nolimits_{t=1}^m {{\beta _t}{{\cal P}_t}},
\end{equation}    
where $d(\cdot,\cdot)$ is commonly implemented with the KLD~\cite{kullback1951information} or $\alpha$-divergence~\cite{renyi1961measures}, and $\beta _t>0$ the mixing coefficient satisfying $\sum\nolimits_{t=1}^m {{\beta _t}}=1$.
The barycenter-based dissimilarity follows a similar formula, given by 
\begin{equation}
D_{\rm{BC}}^{\beta}\left( {{{{\cal P}}_{1:m}}} \right) = \mathop {\min }\limits_{{\cal \bar P}} \frac{1}{m}\sum\nolimits_{t = 1}^m {{\beta _t}d\left( {{{{\cal P}}_t},{{\cal {\bar P}}}} \right)},
\end{equation}
where the divergence measure $d(\cdot,\cdot)$ denotes the Riemannian distance between each distribution ${\cal{P}}_t$ and the barycenter ${\cal {\bar P}}$ which minimizes the defined dissimilarity.
The $f$-dissimilarity is formulated as 
\begin{equation}
{D_f}\left( {{{{\cal P}}_{1:m}}} \right) = \int {f\left( {{p_1}\left( {\mathbf{x}} \right), \cdots, {p_m}\left( {\mathbf{x}} \right)} \right)}  {d{\mathbf{x}}}, 
\end{equation}
where $f$ is a continuous, convex, and homogeneous function.

In fact, from a broader perspective, $D_{\rm{IR}}$ can be seen as a specific case of $D_{\rm{BC}}$.
Some of the existing affinity measures of multiple distributions can also be viewed as special cases or variants of $f$-dissimilarity.
For example, the generalized Jensen-Shannon divergence~\cite{lin1991divergence} is $f$-dissimilarity with 
\begin{equation}
f = \sum\nolimits_{t=1}^m {{\beta _t}{p_t}\left( {\mathbf{x}} \right)\log \left( {{p_t}\left( {\mathbf{x}} \right)} \right)}  - \left({\sum\nolimits_{t=1}^m {{\beta _t}{p_t}\left( {\mathbf{x}} \right)} }\right) \log \left( {\sum\nolimits_{t=1}^m {{\beta _t}{p_t}\left( {\mathbf{x}} \right)} } \right).
\end{equation}
The Jensen-R\'{e}nyi divergence~\cite{hamza2003jensen} and the Matusita's measure of affinity-based distance~\cite{kapur1983matusita} both can be viewed as variants of $f$-dissimilarity termed as $-\log \left( D_f\left( {{{{\cal P}}_{1:m}}} \right) \right)$ with
\begin{equation}
f_{\rm{JR}} = \sum\nolimits_{t=1}^m {{\beta _t}p_t^\alpha \left( {\mathbf{x}} \right)}  - \left( {\sum\nolimits_{t=1}^m {{\beta _t}{p_t}\left( {\mathbf{x}} \right)} } \right)^\alpha, \;
f_{\rm{MMA}} = \left(\prod\nolimits_{t=1}^m   {{p_t}\left( {\mathbf{x}} \right)} \right)^{\frac{1}{m}}.
\end{equation}

Despite extensive definitions found in the literature, generalized divergence measures still face limitations across all three categories mentioned, resulting in quite limited application in the field of machine learning.
For instance, $D_{\rm{IR}}$ and $D_{\rm{BC}}$ require an optimal generalized mean or barycenter distribution, which can be challenging to obtain. 
Similarly, designing an appropriate $f$-function for $f$-dissimilarity tailored to specific data characteristics is a daunting task. 
Furthermore, the lack of effective and efficient sample-based estimators hinders their application in data-driven practical scenarios.

Can we develop a new generalized divergence measure with a closed-form empirical estimator to facilitate its implementation as a loss function in machine learning tasks?
We will show how the classical CS divergence~\cite{principe2000learning} can be extended to a generalized divergence for multiple distributions, taking inspiration from the generalized H\"older’s inequality \cite{finner1992generalization}.
Additionally, we will derive a corresponding closed-form empirical estimator, making it convenient and suitable for various machine-learning applications.

\subsection{Sample-based Estimation}
Note that the majority of divergence measures are defined in a probability rather than sample space.
Hence, we have to resort to density estimation from samples to address this issue. 
Kernel density estimation (KDE)~\cite{parzen1962estimation} serves as an effective technique and has gained significant popularity in various fields, including statistics data analysis~\cite{scott2015multivariate,silverman2018density} and machine learning~\cite{principe2010information,wkeglarczyk2018kernel,kamalov2020kernel}.
Consider a set of independent and identically distributed samples $\left\{{\mathbf{x}}_i\right\}_{i=1}^{n}$, where each sample ${{\mathbf{x}}_i \in {\mathbb{R}}^d}$ is drawn from a multi-variate distribution $\cal P$ with probability density function $p\left(\mathbf{x}\right)$ at position $\mathbf{x}$, the KDE utilizes a kernel function ${\kappa}(\cdot)$ to smooth the observed data points, thereby generating a continuous estimate of the underlying PDF~\cite{scott2015multivariate}:
\begin{equation}\label{eq:KDE}
    \hat p\left({\mathbf{x}}\right) 
    = \frac{1}{n {\sigma}^d}\sum\limits_{i = 1}^n { {\kappa }\left( \frac{{{\mathbf{x}} - {{\mathbf{x}}_i}}}{\sigma} \right)},
\end{equation}
where ${\kappa}(\cdot)$, typically symmetric and smooth, is centered at each data point, scaled by a bandwidth parameter $\sigma$, and has to obey the following properties~\cite{silverman2018density}: $\kappa \left( s \right) \ge 0,\int {\kappa \left( s \right)ds}  = 1,\int {s\kappa \left( s \right)ds}  = 0,\int {{s^2}\kappa \left( s \right)ds}  = {\mu_{\kappa} }<+\infty$. 

In this study, we seek the direct estimation of GCSD from samples by estimating ${\mathbb E}\left[p(\mathbf{x})\right]$ and its polynomials instead of first estimating $p(\mathbf{x})$ and then computing the divergence.
Given that the empirical GCSD estimator is closely related to the KDE, it is important to clarify whether it is susceptible to the curse of dimensionality, a well-known phenomenon in KDE.
We will provide a sound rationale with the \textbf{Remark}~\ref{remark:kde} in Section~\ref{Estimator}.

\section{Generalized Cauchy-Schwarz Divergence}\label{Methods}
In the following subsections, we begin by presenting a novel approach to defining the generalized divergence measure for multiple distributions.
Subsequently, we proceed to analyze some inherent properties of this measure.
\subsection{Definition}\label{sec:proposal-GCSD}
Our proposal is motivated by the classic CSD \cite{principe2000learning}, which draws its inspiration from CS inequality. 
Actually, a more general option is the generalized H\"older’s inequality \cite{finner1992generalization}:
\begin{equation}
\label{eq:h_inqeuality}
\int {\prod\nolimits_{t = 1}^m {\left| {{p_t}\left( {\mathbf{x}} \right)} \right|} d{\mathbf{x}}}  
\le 
\prod\nolimits_{t = 1}^m {{{\left( {\int {{{\left| {{p_t}\left( {\mathbf{x}} \right)} \right|}^m}d{\mathbf{x}}} } \right)}^{{1 \mathord{\left/{\vphantom {1 m}} \right.\kern-\nulldelimiterspace} m}}}} ,
\end{equation}   
from which we obtain the generalized Cauchy-Schwarz divergence as in Definition \ref{def:gcsd}
\footnote{Nielsen et al. have introduced the H\"{o}lder pseudo-divergences (HPDs)~\cite{nielsen2017holder} which, under specific parameter settings, can also yield the classical CSD. 
While both the HPDs and our defined GCSD draw inspiration from the H\"older's inequality, it is important to note that the HPDs focus on measuring two functionals, whereas the GCSD is primarily developed for multiple distributions.

Kapur et al. also discussed generalized Matusita's measure of affinity (MMA) for multiple distributions, analyzed its properties relying on Holder's inequality \cite{kapur1983matusita}, and proposed to use $d({\mathcal P}_{1:m})=1-\text{MMA}({\mathcal P}_{1:m})$ or $-\log\left(\text{MMA}({\mathcal P}_{1:m})\right)$ as a generalized divergence measure.
However, it is also fundamentally different from our proposed GCSD.}.
\begin{globalDef}\label{def:gcsd}
(The generalized Cauchy-Schwarz divergence.)
Let $\left\{\mathcal{P}_t\right\}_{t=1}^m$ be a finite set of probability distributions defined on $\mathcal{X}\in\mathbb{R}^d$ with $p_t(\mathbf{x})$ denoting the probability density of a data point $\mathbf{x}$ from the $t$-th distribution, then the generalized Cauchy-Schwarz divergence amongst $\left\{\mathcal{P}_t\right\}_{t=1}^m$ is defined as:
    \begin{equation}\label{gcsd_def}                
    {D_{{\rm{GCS}}}}({\mathcal P}_{1:m}) =  - \log \frac{{\int {\prod\nolimits_{t = 1}^m {{p_t}\left( {\mathbf{x}} \right)} d{\mathbf{x}}} }}{{\prod\nolimits_{t = 1}^m {{{\left( {\int {p_t^m\left( {\mathbf{x}} \right)d{\mathbf{x}}} } \right)}^{\frac{1}{m}}}} }}.
    \end{equation}
\end{globalDef}        
This definition of divergence for multiple distributions satisfies the {\emph{non-negativity}}, {\emph{symmetry}}, and {\emph{projective invariance}} properties. 
Proofs for the properties mentioned above are presented in Section~\ref{sec:property_1}$\sim$\ref{sec:property_3}.     
\subsection{Non-negativity}\label{sec:property_1}
Given the probability densities $p_t(\mathbf{x}) \ge 0, t=1,\cdots,m$, we can apply the generalized H\"{o}lder's inequality as stated in Eq.~\eqref{eq:h_inqeuality} to obtain the following inequality:
\begin{equation}\label{eq:proof_1_2_1}
0 \le \frac{{\int {\prod\nolimits_{t = 1}^m {{p_i}\left( {\mathbf{x}} \right)} d{\mathbf{x}}} }}{{\prod\nolimits_{t = 1}^m {{{\left( {\int {p_t^m\left( {\mathbf{x}} \right)d{\mathbf{x}}} } \right)}^{\frac{1}{m}}}} }} = \frac{{\int {\prod\nolimits_{t = 1}^m {\left| {{p_t}\left( {\mathbf{x}} \right)} \right|} d{\mathbf{x}}} }}{{\prod\nolimits_{t = 1}^m {{{\left| {\int {p_t^m\left( {\mathbf{x}} \right)d{\mathbf{x}}} } \right|}^{\frac{1}{m}}}} }} \le 1 ,
\end{equation}
and the right part of the Eq.~\eqref{eq:proof_1_2_1} holds equal iff there exist constants $\beta _i > 0$, $i=1,\cdots,m$, such that $\beta _1 p_1 = \beta _2 p_2=\cdots = \beta _m p_m$.
This implies that
\begin{equation}
D_{\rm{GCS}}({\mathcal P}_{1:m}) =  - \log \frac{{\int {\prod\nolimits_{t = 1}^m {{p_t}\left({\mathbf{x}}\right)} d\mathbf{x}} }}{{\prod\nolimits_{t = 1}^m {{{\left( {\int {p_t^m\left({\mathbf{x}}\right)d\mathbf{x}} } \right)}^{\frac{1}{m}}}} }} \ge 0
\end{equation}
and the equation holds iff $\beta _1\mathcal{P}_1 = \beta _2\mathcal{P}_2 = \cdots = \beta _m \mathcal{P}_m$.
That completes the proof.
\subsection{Symmetry}\label{sec:property_2}
By the commutative property of multiplication, it is asserted that 
\begin{equation}
\begin{array}{l}
{D_{\rm{GCS}}}({\mathcal{P}_1}, \cdots ,{\mathcal{P}_i}, \cdots ,{\mathcal{P}_j}, \cdots ,{\mathcal{P}_m})\\
    = {D_{\rm{GCS}}}({\mathcal{P}_1}, \cdots ,{\mathcal{P}_j}, \cdots ,{\mathcal{P}_i}, \cdots ,{\mathcal{P}_m})\\
= - \log \int {\prod\nolimits_{t = 1}^m {{p_t}\left({\mathbf{x}}\right)} d\mathbf{x}}  + \frac{1}{m}\log \prod\nolimits_{t = 1}^m {\left( {\int {p_t^m\left({\mathbf{x}}\right)d\mathbf{x}} } \right)} .
\end{array}
\end{equation}
That completes the proof.

The properties of \emph{Non-negativity}, \emph{Identity}, and \emph{Symmetry} collectively ensure that our proposed GCSD is a valid divergence measure. 
\subsection{Projective Invariance}\label{sec:property_3}
The property of \emph{Projective Invariance} plays a pivotal role in guaranteeing the stability and consistency of quantifying dissimilarity among distributions, regardless of the scales employed to measure density. 
This property primarily emphasizes the geometric structures inherent in these distributions.

Let us consider a collection of positive scalars $\{\beta_t\}_{t=1}^{m}$ and the corresponding distributions $\{\beta_t{\cal P}_t\}_{t=1}^m$.
By applying Definition \ref{def:gcsd}, we can compute the GCSD amongst $\{\beta_t{\cal P}_t\}_{t=1}^m$ as follows:
\begin{equation}\label{eq:eq_beta_gcsd}
    {D_{\rm{GCS}}}\left( \beta_1{\cal P}_1,\cdots,\beta_m{\cal P}_m \right) 
    = - \log \frac{{\int {\prod\nolimits_{t = 1}^m {{\beta _t}{p_t}\left( {\mathbf{x}} \right)} d{\mathbf{x}}} }}{{\prod\nolimits_{t = 1}^m {{{\left( {\int \left({\beta _t}p_t\left( {\mathbf{x}} \right)\right)^md{\mathbf{x}}} \right)}^{\frac{1}{m}}}} }}        .
\end{equation}
Then, it can be deduced that
\begin{equation}\label{eq:proj_invariant}
    \begin{array}{*{20}{l}}
    {D_{{\rm{GCS}}}}\left( \beta_1{\cal P}_1,\cdots,\beta_m{\cal P}_m \right) \\
     =  - \log \int {\prod\nolimits_{t = 1}^m {{\beta _t}{p_t}\left( \mathbf{x} \right)} d\mathbf{x}}  + \frac{1}{m}\sum\nolimits_{t = 1}^m {\log \int {{{\left( {{\beta _t}{p_t}\left( \mathbf{x} \right)} \right)}^m}d\mathbf{x}} } \\
     =  - \log \prod\nolimits_{t = 1}^m {{\beta _t}} \int {\prod\nolimits_{t = 1}^m {{p_t}\left( \mathbf{x} \right)} d\mathbf{x}}  + \frac{1}{m}\sum\nolimits_{t = 1}^m {\log \beta _t^m\int {p_t^m\left( \mathbf{x} \right)d\mathbf{x}} } \\
     =  - \log \int {\prod\nolimits_{t = 1}^m {{p_t}\left( \mathbf{x} \right)} d\mathbf{x}}  + \log \prod\nolimits_{t = 1}^m {{{\left( {\int {p_t^m\left( \mathbf{x} \right)d\mathbf{x}} } \right)}^{\frac{1}{m}}}}. \\
     \end{array}
\end{equation}
The final expression in Eq.~\eqref{eq:proj_invariant} is equivalent to the definition of $D_{\rm{GCS}}({\cal P}_{1:m})$ given in Definition \ref{def:gcsd}.
This demonstrates the projective invariance of the proposed GCSD:
\begin{equation}
    {D_{{\rm{GCS}}}}\left({\beta _1}{{\cal P}_1}, \cdots, {\beta _m}{{\cal P}_m} \right) = D_{\rm{GCS}}({\cal P}_{1:m}).
\end{equation}
    
\section{Empirical Estimation and Power Test}\label{Analysis}  
    \subsection{Sample Estimator}\label{Estimator}  
    Based on the kernel estimation~\cite{parzen1962estimation}, we derive a nonparametric estimator for our GCSD that eliminates the need for prior knowledge of the underlying probability distribution, making it more versatile and applicable in data-driven machine-learning tasks.

    Rewrite the definition of GCSD in Eq.~\eqref{gcsd_def} as 
    \begin{equation}\label{eq:split_gcsd}
    \begin{array}{*{20}{l}}
    {{D_{{\rm{GCS}}}}({\cal P}_{1:m}) =  - \log \underbrace {\int {\left( {\prod\nolimits_{t = 1}^m {{p_t}\left( {\mathbf{x}} \right)} } \right)d{\mathbf{x}}} }_{{V_1}} + \frac{1}{m}\sum\nolimits_{t = 1}^m {\log \underbrace {\int {p_t^m\left( {\mathbf{x}} \right)d{\mathbf{x}}} }_{{V_2}}} }.
    \end{array}  
    \end{equation}
    Then, $V_1$ in Eq.~\eqref{eq:split_gcsd} can be approximated by taking expectation over a specific distribution ${\cal P}_t$ and averaging the expectations over all distributions:
    \begin{equation}
    V_1  =  {\frac{1}{m}\sum\nolimits_{t = 1}^m {{{\mathbb E}_{{\cal P}_t}}\left[\prod\nolimits_{k \ne t}^m {{p_k}\left( X \right)} \right]} },
    \end{equation}
    where the outer-stage expectation aims to maintain symmetry. 
    One can also randomly select from the overall set a distribution as ${\cal P}_t$ and omit the outer-stage expectation to implement the estimation.
    
    Let ${\mathbf{x}}_i^t \in {\mathbb{R}}^d$ represents the $i$-th sample drawn from the $t$-th distribution ${\cal{P}}_t$, where $1\leq i \leq n_t$ and $n_t$ is the number of samples. 
    Then, with ${\kappa _\sigma }(\cdot)$ implemented by the Gaussian kernel
    \begin{equation}
    {\kappa _\sigma }\left( {\bf{x}} \right) = \frac{1}{{{{\left( {\sqrt {2\pi } \sigma } \right)}^{d}}}}\exp \left( { - \frac{{\left\| {\bf{x}} \right\|_2^2}}{{2{\sigma ^2}}}} \right),        
    \end{equation}
    ${\hat V}_1$ is estimated as
    \begin{equation}\label{eq:cross_entropy_gcsd}
    \begin{array}{l}
    % \circ{\scriptsize{1}}
    {\hat V}_1 \approx  {\frac{1}{m}\sum\nolimits_{t = 1}^m {\frac{1}{{{n_t}}}\sum\nolimits_{j = 1}^{{n_t}} {\left( {\prod\nolimits_{k \ne t}^m {\frac{1}{{{n_k}}}\sum\nolimits_{i = 1}^{{n_k}} {{\kappa _\sigma }\left( {{\mathbf{x}}_j^t - {\mathbf{x}}_i^k} \right)} } } \right)} } } \\    
    \end{array}.
    \end{equation}
    $V_2$ in Eq.~\eqref{eq:split_gcsd} can be estimated similarly as below:
    \begin{equation}\label{eq:power_entropy_gcsd}
    \begin{array}{*{20}{l}}
    {{V_2} = {{\mathbb E}_{{\cal P}{_t}}}\left[ {p_t^{m - 1}\left( X \right)} \right]} \Rightarrow \\ 
    {{\hat V}_2 \approx \frac{1}{{{n_t}}}\sum\nolimits_{j = 1}^{{n_t}} {p_t^{m - 1}\left( {{{\mathbf{x}}_j}} \right)} }
    {\approx \frac{1}{{{n_t}}}\sum\nolimits_{j = 1}^{{n_t}} {{{\left( {\frac{1}{{{n_t}}}\sum\nolimits_{i = 1}^{{n_t}} {{\kappa _\sigma }\left( {{\mathbf{x}}_j^t - {\mathbf{x}}_i^t} \right)} } \right)}^{m - 1}}} .}
    \end{array}
    \end{equation}
    Take Eq.~\eqref{eq:cross_entropy_gcsd} and ~\eqref{eq:power_entropy_gcsd} into Eq.~\eqref{eq:split_gcsd}, we obtain:
    \begin{equation} \label{eq:estimator_GCSD} 
    \begin{aligned}            
    {{\hat D}_{{\rm{GCS}}}({\mathcal{P}_{1:m}})} \approx &- \log \left( {\frac{1}{m}\sum\nolimits_{t = 1}^m {\frac{1}{{{n_t}}}\sum\nolimits_{j = 1}^{{n_t}} {\prod\nolimits_{k \ne t}^m {\frac{1}{{{n_k}}}\sum\nolimits_{i = 1}^{{n_k}} {{\kappa _\sigma }\left( {{\mathbf{x}}_j^t - {\mathbf{x}}_i^k} \right)} } } } } \right) \\
    &+ \frac{1}{m}\sum\nolimits_{t = 1}^m {\log \frac{1}{{{n_t}}}\sum\nolimits_{j = 1}^{{n_t}} {{{\left( {\frac{1}{{{n_t}}}\sum\nolimits_{i = 1}^{{n_t}} {{\kappa _\sigma }\left( {{\mathbf{x}}_j^t - {\mathbf{x}}_i^t} \right)} } \right)}^{m - 1}}} }.
    \end{aligned}  
    \end{equation} 
    It is important to note that the above estimator is differentiable and well-suited for application in machine learning.      	

    \begin{globalRem}\label{remark:kde}
    (The kernel-based sample estimator ${\hat D}_{{\rm{GCS}}}$ can effectively deal with features learned from a deep-learning network.)
    We invoke the KDE or more accurately, Parzen windowing, not to estimate full density functions, but to capture the underlying properties of the density and fulfill the estimation of the high-level information–theoretic quantity, GCSD. 
    Validation on synthetic data has demonstrated that ${\hat D}_{{\rm{GCS}}}$ is effective and robust in addressing high-dimensional features, as illustrated in Figure~\ref{fig:power_test}. 
    Additionally, we propose incorporating GCSD in deep learning to measure divergence in the latent space, especially the dimension-reduced bottleneck, thus greatly alleviating the challenges posed by the curse of dimensionality if it exists. 
    % Moreover, there is abundant indirect yet compelling evidence to substantiate the effectiveness of kernel-based measures in handling data across various deep-learning applications.
    % This is supported by the successful utilization of kernel-based measures such as the CSD~\cite{yu2023cauchy}, MMD~\cite{smola2006maximum,gretton2012kernel}, and HSIC~\cite{gretton2005measuring}
    % \footnote{In their article~\cite{yu2023cauchy}, Yu et al. have demonstrated the inherent connection between CSD estimator and the MMD, as well as between the KDE-based CS-QMI and the HSIC.}.    
    \end{globalRem}    
    
    \subsection{Bias Analysis}
    The estimator ${\hat D}_{\rm{GCS}}(\mathcal{P}_{1:m})$ in Eq.~\eqref{eq:estimator_GCSD} relies on two key components, ${\hat V}_1$ and ${\hat V}_2$ in Eqs.~\eqref{eq:cross_entropy_gcsd}$\sim$~\eqref{eq:power_entropy_gcsd}.
    We focus on conducting a preliminary analysis of the bias related to ${\hat V}_1$ and ${\hat V}_2$ for simplicity, and derive that they are both unbiased estimators under certain conditions.
    
    \textbf{Bias of ${\hat V}_1$. }
    Take expectation on ${\hat V}_1$ in Eq.~\eqref{eq:cross_entropy_gcsd}, it is derived that
    \begin{equation}
        \begin{array}{l}
        {\mathbb E}\left[ {{{\hat V}_1}} \right] = {\mathbb E}\left[ {\frac{1}{m}\sum\nolimits_{t = 1}^m {\frac{1}{n}\sum\nolimits_{i = 1}^n {\prod\nolimits_{k \ne t}^m {\frac{1}{n}\sum\nolimits_{j = 1}^n {{\kappa _\sigma }\left( {{\mathbf{x}}_i^k - {\mathbf{x}}_j^t} \right)} } } } } \right]\\
         = {\mathbb E}\left[ {\prod\nolimits_{k \ne t}^m {{\kappa _\sigma }\left( {{{\mathbf{x}}^k} - {{\mathbf{x}}^t}} \right)} } \right]  
         \quad subscripts\; are\; omitted\; for\; simplicity.\\
         = \int {{p_t}\left( {{{\mathbf{x}}^t}} \right)d{{\mathbf{x}}^t}\prod\nolimits_{k \ne t}^m {{\kappa _\sigma }\left( {{{\mathbf{x}}^k} - {{\mathbf{x}}^t}} \right){p_k}\left( {{{\mathbf{x}}^k}} \right)d{{\mathbf{x}}^k}} } 
         \quad Let\; {s^k} = \frac{{{{\mathbf{x}}^k} - {{\mathbf{x}}^t}}}{\sigma },\\
         = \int {{p_t}\left( {{{\mathbf{x}}^t}} \right)d{{\mathbf{x}}^t}\prod\nolimits_{k \ne t}^m {\kappa \left( {{{\mathbf{s}}^k}} \right){p_k}\left( {{{\mathbf{x}}^t} + \sigma {{\mathbf{s}}^k}} \right)d{{\mathbf{s}}^k}} } 
         \quad Let\;\dot p_k, \ddot p_k\; be\; 1st,\; 2nd\; derivative\\
         = \int {{p_t}\left( {{{\mathbf{x}}^t}} \right)d{{\mathbf{x}}^t}\prod\nolimits_{k \ne t}^m {( {{p_k}\left( {{{\mathbf{x}}^t}} \right) + \sigma {{\mathbf{s}}^k}{{{\dot p}}_k}\left( {{{\mathbf{x}}^t}} \right) + \frac{1}{2}{\sigma ^2}{{\left( {{{\mathbf{s}}^k}} \right)}^2}{{{\ddot p}}_k}\left( {{{\mathbf{x}}^t}} \right) + {\cal O}\left( {{\sigma ^2}} \right)} )\kappa \left( {{{\mathbf{s}}^k}} \right)d{{\mathbf{s}}^k}} }.
         % = \int {{p_t}\left( {{x^t}} \right)\prod\nolimits_{k \ne t}^m {\left( {{p_k}\left( {{x^t}} \right) + \frac{1}{2}{\sigma ^2}{{{\ddot p}}_k}\left( {{x^t}} \right){\kappa_2 } + {\cal O}\left( {{\sigma ^2}} \right)} \right)} } d{x^t}
         % \quad Let\; {\kappa_2 } = \int {{s^2}\kappa \left( s \right)ds}
        \end{array}
    \end{equation}
    Kernel function $\kappa$ satisfies $\int {\kappa \left( s \right)ds}  = 1$, $\int {s\kappa \left( s \right)ds}  = 0$, and $0<\int {{s^2}\kappa \left( s \right)ds}  = {\kappa_2 }<+\infty$~\cite{wasserman2006all}, thus we obtain that
    \begin{equation}
        \begin{array}{l}
        {\rm{bias}}\left[ {{{\hat V}_1}} \right] = {\mathbb E}\left[ {{{\hat V}_1}} \right] - {{V}_1}\\
         = \int {{p_t}\left( \mathbf{x} \right)\prod\nolimits_{k \ne t}^m {\left( {{p_k}\left( x \right) + \frac{1}{2}{\sigma ^2}\kappa_2{{{\ddot p}}_k}\left( x \right) + {\cal O}\left( {{\sigma ^2}} \right)} \right)d\mathbf{x}} }  - \int {\left( {\prod\nolimits_{t=1}^m {{p_t}\left( \mathbf{x} \right)} } \right)d\mathbf{x}} 
        \end{array}
    \end{equation}
    and $\mathop {\lim }\limits_{\scriptstyle\sigma  \to 0\hfill} bias\left[ {{{\hat V}_1}} \right] = 0$ with ${{\ddot p}_k} <  + \infty $.
    
    In conclusion, when $\sigma \to 0$ and the second-order derivatives of all ${p_t}$s are bounded, ${\hat V}_1$ can be regarded as an unbiased estimation of ${V}_1$.

    \textbf{Bias of ${\hat V}_2$. }
    Bias of ${\hat V}_2$ can be derived similarly to that of the ${\hat V}_1$ as follows:
    \small{
    \begin{equation}
        \begin{array}{l}
        {\mathbb E}\left[ {{{\hat V}_2}} \right]
         % = {\mathbb E}\left[ {\frac{1}{{{n_t}}}\sum\nolimits_{j = 1}^{{n_t}} {{{\left( {\frac{1}{{{n_t}}}\sum\nolimits_{i = 1}^{{n_t}} {{\kappa _\sigma }\left( {{\bf{x}}_j^t - {\bf{x}}_i^t} \right)} } \right)}^{m - 1}}} } \right]
         = {\mathbb E}\left[ {\kappa _{^\sigma }^{m - 1}\left( {{\bf{x}}_j^t - {\bf{x}}_i^t} \right)} \right]\\
         = \int {{p_t}\left( {\bf{x}} \right){{\left( \int{\left[ {{p_t}\left( {\bf{x}} \right) + \sigma s{{\dot p}_t}\left( {\bf{x}} \right) + \frac{1}{2}{\sigma ^2}{s^2}{{\ddot p}_t}\left( {\bf{x}} \right) + O\left( {{\sigma ^2}} \right)} \right]\kappa \left( s \right)ds} \right)}^{m - 1}}d{\bf{x}}} \\
         \approx \int {p_t^m\left( {\bf{x}} \right)d{\bf{x}}} 
           \leftarrow {\left\{ 
           \begin{array}{l}
            \int {{s^2}\kappa \left( s \right)ds}  <  + \infty ,\int {s\kappa \left( s \right)ds}  = 0,\\
            \int {\kappa \left( s \right)ds}  = 1,{{\ddot p}_k} <  + \infty ,\sigma  \to 0.
            \end{array} \right.}
        \end{array}
    \end{equation}
    }
    It is concluded that ${\hat V}_2$ is an unbiased estimation of ${V}_2$ under the same assumptions or conditions as for ${\hat V}_1$.
    
\subsection{Complexity Analysis} \label{sec:complexity}       
\begin{figure}[tbp]
        \begin{minipage}{0.45\linewidth}
            \centering
            \resizebox{\linewidth}{!}{%
                \begin{tabular}{ll}    \toprule
                    Methods & Complexity                  \\ \midrule
                    pKLD    & ${\cal{O}}((6d+11)m^2n^2)$  \\
                    PMMD    & ${\cal{O}}((6d+6)m^2n^2)$    \\
                    pCSD    & ${\cal{O}}((6d+6)m^2n^2)$    \\
                    GCSD    & ${\cal{O}}((2d+3)m^2n^2)$ \\ \bottomrule
                \end{tabular} 
            }
            \captionsetup{type=table}
            \caption{Complexity statistics}
            \label{tab:complexity}
        \end{minipage}%
        \hfill
        \begin{minipage}{0.43\linewidth}
            \centering
            \includegraphics[width=1.0\linewidth]{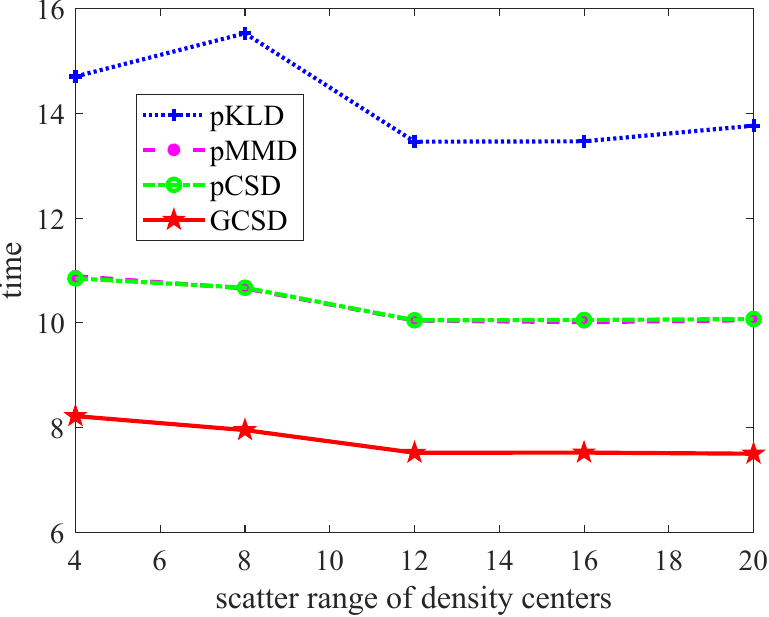}
            \caption{Run time on synthetic data}
            \label{fig:time_consumption}
        \end{minipage}
    \end{figure}

    Consider a collection of $i.i.d.$ datasets $\left\{X_i\right\}_{i=1}^{m} \in \mathbb{R}^d$ drawn from $m$ distributions, each with the same sample size $n$.
    The computation complexity of a generalized divergence measure is determined by the total number of mathematical operations involved.
    For instance, when computing ${\hat D}_{\rm{GCS}}$, the process involves $(d+1)m^2n^2+m(n+1)$ additions or subtractions, $dm^2n^2+2mn(m-1)$ multiplications, $mn(m+n)+2m+1$ divisions, $mn^2(2m-1)$ exponentiations, and $2m+1$ logarithms. 
    This results in a total of $2(d+1)m^2n^2+3m^2-mn+5m+2$ mathematical operations, yielding a complexity of ${\cal{O}}(2(d$+$1)m^2n^2)$. 
    We have chosen several highly representative average pairwise divergences, the average-pairwise KLD, MMD, and CSD (pKLD, pMMD, and pCSD), as comparative methods, and the results are summarized in Table \ref{tab:complexity}. 
	
    Please note that our proposed GCSD requires significantly fewer operations compared to pKLD, pMMD, and pCSD, with at most about one-third of the operations needed. 
    It is important to mention that the computational complexity of all the comparison methods is of the same order when considering the distribution and sample size. 
    The difference lies in the multiples relative to the dimensionality, $d$, of the data features. 
    
    The running time of various metrics on the synthetic data described in the following section, as depicted in Figure~\ref{fig:time_consumption}, can provide support for our complexity analysis.
    \subsection{Power Test}  
    \subsubsection{Data Preparation}\label{sec:sata_preparation}
    We employ a set of multiple distributions to generate synthetic datasets, facilitating the assessment of our GCSD in comparison with three average-pairwise competitors termed as pKLD, pMMD, and pCSD. 
    It consists of ten distributions: three Gaussian, three uniform, two bimodal Gaussian, and two bimodal uniform distributions. 
    Further details and visualization of the synthetic data can be found in \ref{app:powertest}.
    
    In this study, we measure the dispersion of the overall set by calculating the distance between the density centers of the two furthest sub-distributions among the five.
    More specifically, we randomly generate 1000 samples for each sub-distribution, resulting in a total of 10000 samples to construct the synthetic dataset for the test.
    Visualization of the synthetic data with scatter range $r=20$ (The distance from the leftmost density center to the rightmost density center) is shown in Figure~\ref{fig:syn_distributions}.  
    To evaluate the effectiveness and robustness of the proposed measure, we generate five data sets with increasing scaling ranges $r\in\left\{4,8,12,16,20\right\}$.

    \subsubsection{Effectiveness Test} %And Discriminability
\begin{figure}[tbp]
        \vspace{-1cm}
        \centering
        \subfloat[With varying scatter range.]{
            \includegraphics[width=0.45\linewidth]{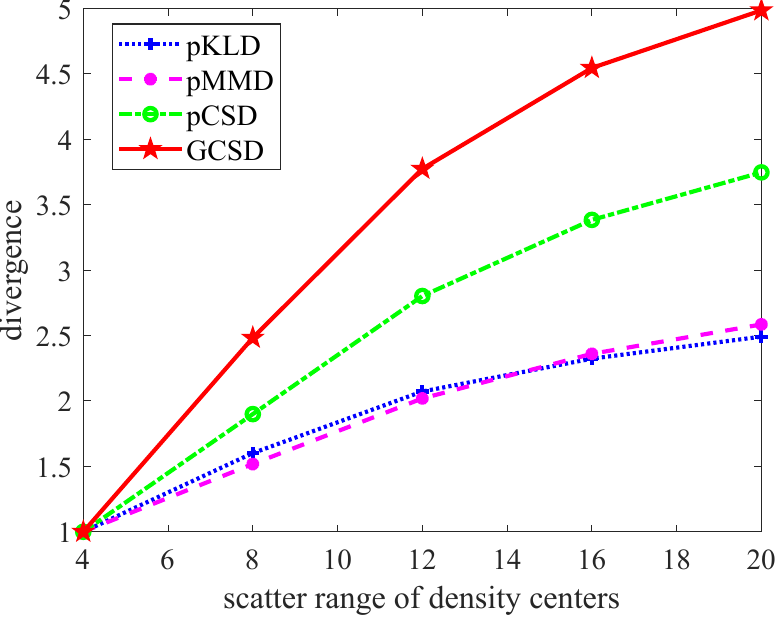}
            \label{fig:div_with_scatter_range}}
        \hfill
        \subfloat[With varying dimension.]{
            \includegraphics[width=0.47\linewidth]{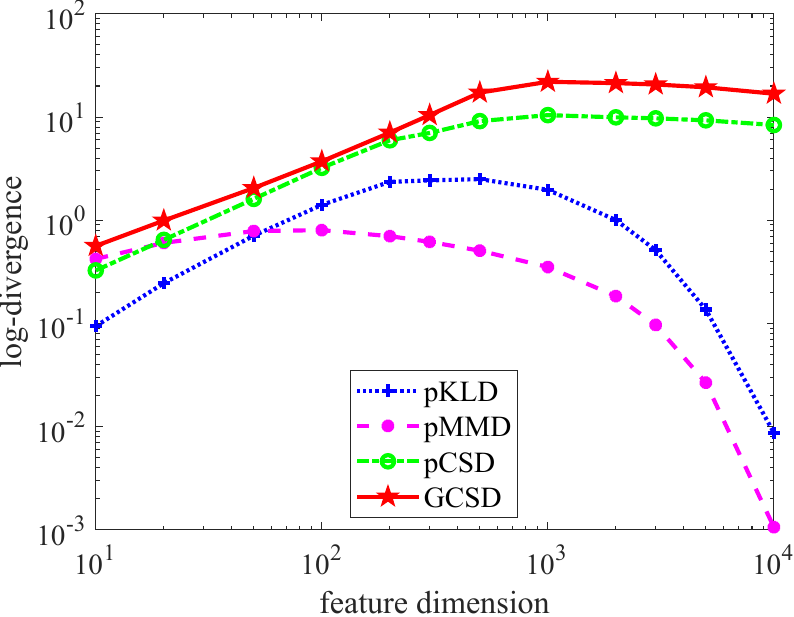}
            \label{fig:div_with_dim}}
        \caption{Comparison of divergence measures. 
            (a) The evaluation is conducted on univariate random data samples from 10 distributions. 
            To ensure a fair comparison, measurements of all the metrics are normalized by dividing them by their respective minimum values. 
            (b) The evaluation is conducted on multivariate samples from 3 distributions (${\cal{N}}_1,{\cal{N}}_2$, and ${\cal{U}}_1$ as illustrated in \ref{sec:sata_preparation} and Figure~\ref{fig:syn_distributions}) with their dimension varying from $10^1$ to $10^4$.
            To provide a clear visual representation, we use logarithmic scaling for all measured values.}
        \label{fig:power_test}
    \end{figure}

    Intuitively, a larger scaling range in our designed distribution set should yield more dispersed distributions, thereby increasing the divergence value. 
    The results of the tests, depicted in Figure~\ref{fig:div_with_scatter_range}, indicate that the proposed GCSD and the three comparative average-pairwise divergence measures show a synchronized trend with increasing scaling range $r$, thus emphasizing their effectiveness in quantifying the total divergence amongst multiple distributions.
    These experiments were conducted with 10 Monte Carlo runs for each setting using the designed synthetic datasets, and the results were averaged to ensure a reliable assessment.
    
    \subsubsection{High-dimensional Robustness Test}
    To further evaluate the performances of GCSD in high-dimensional space, we conduct experiments on multivariate data across varying dimensions.
    The feature dimensions in each sub-distribution are varied from $10^1$ to $10^4$ and the results are presented in Figure~\ref{fig:div_with_dim}. 
    As can be discovered, the proposed GCSD exhibits an ascending trend with the feature dimension and then a minor performance degradation when the feature dimension of testing samples exceeds $10^3$. 
    In contrast, the pKLD and pMMD tend to struggle or even fail to measure the divergence of data with high dimensions. 
	
\section{Application for Deep Clustering}\label{sec:clustering} 
    \subsection {Advancing Clustering}\label{sec:extension_clustering}      
    Given a dataset $X$ with $n\; i.i.d.$ samples $\left\{\mathbf{x}_i\right\}_{i=1}^n$, the task of (hard) clustering refers to dividing the $n$ samples into $m$ distinct clusters $C_1, C_2,\cdots, C_m$, such that 
    \begin{equation}
    \begin{split}            
    \left\{C_t\right\}_{t=1}^m=\arg \mathop {\max }\limits_{{C_t} \subset X} 
    {\rm{Dis}}\left( {{C_1}, \cdots ,{C_m}} \right) 
    \quad s.t.\quad\bigcup\nolimits_{t = 1}^m {{C_t}}  = X.
    \end{split}
    \end{equation}
    where ``Dis" refers to a dissimilarity measure.
    
    A lingering question, however, is how to partition the entire observation set $X$ into $m$ groups. 
    A practical approach is to learn the cluster assignments, represented as a soft assignment matrix $A=\left[a_{i,j}\right] \in \mathbb{R}^{n \times m}$, directly from the data using a network. 
    Each row in $A$ corresponds to a soft pseudo-label $y_i$=$\left[ {{a_{i,1}},{a_{i,2}}, \cdots ,{a_{i,m}}} \right]$, and satisfies $\sum\nolimits_{t=1}^m {{a_{i,t}}}=1$, where $a_{i,t}$ represents the crisp cluster assignment of data sample $i$ belonging to cluster $C_t$. 
    The proposed GCSD measure can then be extended for clustering by leveraging the assignment matrix $A$.
    \begin{globalProp}\label{prop1}
        Given dataset $X=\left\{\mathbf{x}_i\right\}_{i=1}^n$ with $\mathbf{x}_i \in \mathbb{R}^d$ and its cluster-assignment matrix $A\in \mathbb{R}^{n \times m}$, the generalized Cauchy-Schwarz divergence amongst the clusters that have been partitioned from $X$ with the assignment matrix $A$ can be computed as~\footnote{Further details of the deduction can be found in \ref{app:proof}.}:
        \begin{equation}\label{eq:cluster_GCSD}
        \begin{aligned}
        {\hat D}_{\rm{GCS}}^{A} ({\cal P}_{1:m}) =  
        &- \log \left( {\frac{1}{m}{\rm{sum}}\left( {\frac{{{{\left({A^T}\right)}^{m - 1}}}}{{KA}}{\rm{prod}}\left( {KA} \right)} \right)} \right) \\
        &+ \frac{1}{m}{\rm{tr}}\left( {\log \left( {{{\left( {{A^T}} \right)}^{m - 1}}{{\left( {KA} \right)}^{m - 1}}} \right)} \right),
        \end{aligned}
        \end{equation}
        where $K$ represents the Gram matrix obtained by evaluating the positive definite kernel ${\kappa _\sigma}$ on all sample pairs, such that $K_{i,j} = {\kappa _\sigma }({\bf x}_i-{\bf x}_j)$. 
        The notation ${\rm{sum}}(\cdot)$ signifies the summation of all elements within a matrix,
        ${\rm{prod}}(A)$ calculates the product of row elements in matrix $A$, yielding a column vector.
        For instance, ${\rm{prod}}(A)=\left[ {\prod\nolimits_{j = 1}^m {{a_{1,j}}} , \cdots ,\prod\nolimits_{j = 1}^m {{a_{n,j}}} }\right]^T$. 
        Furthermore, ${\rm{tr}}(\cdot)$ denotes the trace of the provided matrix. 
        Additionally, the symbols $A^{n}$, and $\frac{A}{B}$ represent element-wise exponentiation, and division for matrices, respectively.
    \end{globalProp}  
    
    \subsection{Deep Divergence-based Clustering Framework}
    The deep embedded clustering (DEC) method \cite{xie2016dec} significantly improves clustering performance on high-dimensional complex data by employing neural networks to extract features and optimizing a KLD-based clustering objective with a self-training target distribution. 
    Subsequently, research on deep learning-based clustering methods has experienced a substantial surge \cite{guo2017improved,enguehard2019semi,ren2019semi}. 
    In recent years, Jenssen et al. adeptly integrated deep learning with divergence maximization to introduce a novel clustering framework, termed deep divergence-based clustering (DDC) \cite{kampffmeyer2019deep,trosten2021reconsidering,trosten2019recurrent}, effectively harnessing the potential of divergence measures for clustering applications.   
    It should be noted that Jenssen et al. utilized the pCSD in their DDC methods, which is not a generalized divergence measure.
    
    \begin{figure}[tbp]
        % \vspace{-1cm}
        \centering
        \includegraphics[width=0.70\linewidth]{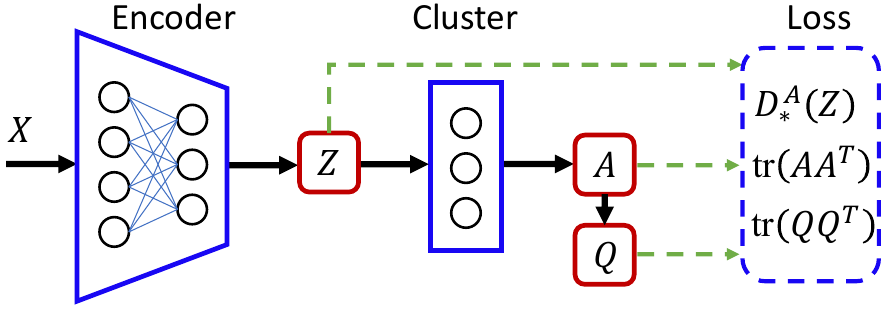}
        \caption{Deep divergence-based clustering framework.
            The Encoder can be implemented using various neural network architectures, such as a simple multilayer perceptron (MLP) for flattened data, a convolutional neural network (CNN) for image-like two-dimensional data, or recurrent neural networks (RNNs) like GRU or LSTM for sequential data.
            The Cluster can be implemented using a fully connected layer followed by softmax activation, allowing for the creation of a cluster assignment matrix.
            The loss function $\rm{Loss}$ integrates a generalized divergence measure $D_*^A(\cdot)$ and two regularization terms on assignment matrix $A$ to preserve its simplex property.}
        \label{fig:framework}
    \end{figure}    
    Figure~\ref{fig:framework} provides a conceptual overview of the deep divergence-based clustering framework. 
    First, the input samples $\{X_i\}_{i=1}^{n}$ to be clustered are mapped to latent representations $\{Z_i\}_{i=1}^{n}$ using a deep neural network, denoted as the Encoder. 
    Subsequently, these latent representations are passed through the Cluster network, which consists of a fully connected output layer with a softmax activation function. 
    This process produces the (soft) cluster membership vectors $\{{\mathbf{\alpha}}_{i}\}_{i=1}^{n}$ and consequently constructs the assignment matrix $A$. 
    
    Ultimately, the framework can be trained end-to-end by maximizing the generalized divergence among the latent groups virtually clustered by the assignment matrix $A$:
    \begin{equation}
    \label{eq:objective1}
    \theta  = \arg \mathop {\max }\limits_{\theta \in {\Theta}} {{\hat D}_{*}^A}\left( {Z} \right).
    \end{equation}
    Here, ${\hat D}_{*}^A$ refers to either the GCSD in Eq.~\eqref{eq:cluster_GCSD} or another kind of generalized divergence measure.
    
    To optimally cluster the data, the assignment matrix $A$ requires appropriate regularization. 
    Following \cite{kampffmeyer2019deep}, we add two terms to the objective function in Eq.~\eqref{eq:objective1} to form the total loss function:
    \begin{equation}\label{eq:lossfunc}
    L={D_{*}^{A}}\left( {Z} \right) + {\lambda _2}{\rm{tr}}\left( {{AA^T}} \right) + {\lambda _3}{\rm{tr}}\left( {Q{Q^T}} \right),
    \end{equation}
    where ${\lambda _2}>0,{\lambda _3}>0$ are the weights used for regularization.
    
    The second term in the loss function is calculated as the trace of the matrix $AA^T$, which promotes orthogonality among the clusters in the $m$-dimensional observation space. 
    The third term enforces the proximity of the cluster membership vectors to a corner of the simplex. 
    In this paper, $Q \in \mathbb{R}^{n \times m}$ has the same dimension as $A$ and its $(i,j)$-th entry is formulated as:
    \begin{equation}\label{eq:l3}
    {Q_{i,j}} = \exp \left( { - {{\left| {{{\boldsymbol{\alpha}} _i} - {{\mathbf{e}}_j}} \right|}^2}} \right),
    \end{equation}
    where ${\boldsymbol{\alpha}}_i$ is the $i$-th row of matrix $A$ and ${\mathbf{e}}_j \in \mathbb{R}^m$ is a unit vector representing the $j$-th corner of the simplex. 
    By employing this regularizer, we ensure that the model generates a simplex assignment matrix $A$.
        
    \subsection{Clustering Experiment}\label{Experi}\label{sec:evaluation}
    We evaluate the proposed divergence measure through clustering experiments conducted on several image and time-series datasets. 
    The comparison methods include three classical baselines: K-means~\cite{hartigan1979algorithm}, spectral clustering (SC)~\cite{shi2000normalized}, and DEC~\cite{xie2016dec}, the SOTA deep divergence-based clustering approach (DDC)~\cite{kampffmeyer2019deep} and two recent proposed deep clustering methods EDESC~\cite{cai2022efficient} and VMM~\cite{stirn2024vampprior}. 
    Please notice that the DDC employs the average-pairwise CSD in the loss function~\cite{kampffmeyer2019deep}.
    All models are trained for 100 epochs using a learning rate of $1e^{-3}$ and a batch size of 100. 
    The weights of the regularization term in the loss function are set to $\lambda_2=\lambda_3=0.5$.
        
    \subsubsection{Evaluate metrics}
    We use two different metrics to evaluate the partition quality of the trained model. 
    The first metric is the unsupervised clustering accuracy, {\small${ {\rm{ACC}} = \mathop {\max }\limits_{\cal M} \sum\nolimits_i {{{\delta \left( {{l_i} = {\cal M}\left( {{c_i}} \right)} \right)} \mathord{\left/{\vphantom {{\delta \left( {{l_i} = M\left( {{c_i}} \right)} \right)} n}} \right.\kern-\nulldelimiterspace} n}} }$}, where $l_i$ refers to the ground truth label, $c_i$ to the assigned cluster of data point $i$, and $\delta(\cdot)$ is the indicator function. $\cal M$ is the mapping function that corresponds to the optimal one-to-one assignment of clusters to label classes.
    The second evaluation measure is the normalized mutual information (NMI), {\small${ {\rm{NMI}} = {{2I\left( {l,c} \right)} \mathord{\left/
                    {\vphantom {{2I\left( {l,c} \right)} {\left( {H\left( l \right) + H\left( c \right)} \right)}}} \right.\kern-\nulldelimiterspace} {\left( {H\left( l \right) + H\left( c \right)} \right)}}}$},
    where $I(\cdot,\cdot)$ and $H(\cdot)$ denote mutual information and entropy functions, respectively.
    
    \subsubsection{Experiments On Image Data}\label{sec:exp-image}  
    \textbf{Datasets. }
    Clustering experiments are conducted on three image datasets: MNIST \cite{lecun1998gradient}, Fashion \cite{xiao2017fashion}, and STL10 \cite{coates2011analysis}, utilizing deep clustering framework illustrated in Figure~\ref{fig:framework}.   
    
    \textbf{Implementation Details.}
    In particular, for MNIST and Fashion datasets, we employ a two-layer convolutional extractor as the encoder with nodes $32$, $64$, and convolution filter sizes of $5\times5$. 
    Each layer is followed by a 2$\times$2 max pooling and a ReLU activation. 
    For the cluster component, we implement a fully connected (FC) layer with 100 nodes, followed by another ReLU activation and a Softmax layer with a suitable dimensionality for the desired number of clusters. 
    Batch normalization is applied to the FC layer for enhanced stability.
    For the STL10 dataset, we follow \cite{cai2022efficient} to render the Resnet50 to extract the 2048 features. 
    Results for DDC~\cite{kampffmeyer2019deep} and EDESC~\cite{cai2022efficient} on STL10 are also based on features extracted with Resnet50.    

    \textbf{Results.}
    The findings presented in Table~\ref{tab:real-results} clearly indicate that our method exhibits exceptional performance in terms of ACC and NMI. 
    To be more specific, the proposed GCSD method outperforms other methods on FashionMNIST and STL10 datasets and achieves performance near to the state-of-the-art on MNIST dataset.
    \begin{table}[tbp]\small
        \centering
        \caption{Clustering results on image datasets. }
        \label{tab:real-results}
        % \resizebox{\columnwidth}{!}{%
        \begin{tabular}{@{}lcccccc@{}}            
            \toprule
            \multirow{2}{*}{Methods}           & \multicolumn{2}{c}{MNIST}  & \multicolumn{2}{c}{Fashion}       & \multicolumn{2}{c}{STL10}                  \\ %\cmidrule(l){2-7} 
            & ACC   & NMI                & ACC   & NMI                  & ACC  & NMI                 \\ \cmidrule(r){1-7}
            Kmeans~(\cite{hartigan1979algorithm}, 1979)  & 0.53      & 0.50           & 0.51      & 0.51          & 0.22     & 0.14                \\
            SC~(\cite{shi2000normalized}, 2000)     & 0.69      & 0.77          & 0.56      & 0.57          & 0.17     & 0.11                  \\
            DEC~(\cite{xie2016dec}, 2016)    & 0.77     & 0.72         & 0.58      & 0.62         & 0.17     & 0.05                     \\ % MLP
            DDC~(\cite{kampffmeyer2019deep}, 2019)     & 0.85      & 0.79          & 0.68      & 0.61          & 0.84     & 0.80                         \\ 
            EDESC~(\cite{cai2022efficient}, 2022)   & 0.91      & 0.86            & 0.63      & 0.67          & 0.74 & 0.69            \\ 
            VMM~(\cite{stirn2024vampprior}, 2024)   & \textbf{0.96}      & \textbf{0.90}        & 0.71 & 0.65          & --  & --         \\  \midrule
            % SPC~\cite{mahon2021selective}   & \bf{0.99} & \bf{0.97}   & 0.68      & \bf{0.74}     & --  & --                 \\ 
            GCSD    & 0.95 & 0.89 & \textbf{0.72} & \textbf{0.65}   & \textbf{0.92}    & \textbf{0.85}          \\ \bottomrule
        \end{tabular}        
        % }
        \begin{tablenotes}\footnotesize
            \item {
            It is worth noting that DDC~\cite{kampffmeyer2019deep} uses the average-pairwise CSD as the divergence measure.
            }
        \end{tablenotes}
    \end{table}
    
\begin{figure}[tbp]
        \subfloat[GCSD]{\includegraphics[width=0.485\linewidth]{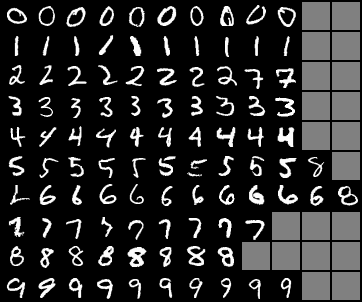}\label{fig:clustered_example}}
        \hfill
        \subfloat[DDC]{\includegraphics[width=0.485\linewidth]{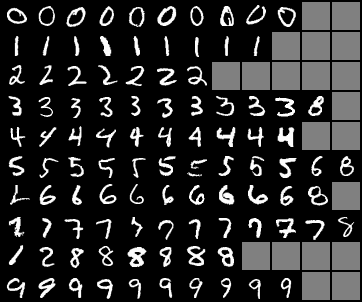}\label{fig:confusion_matrix}}
        \caption{Visualization of the clustered examples.
            A mini-batch of 100 samples clustered into 10 groups, each of which presented in a line.
            Groups with fewer clustered samples are filled with random noises.}
        \label{fig:clustered_examples}
    \end{figure}

    In addition to quantitative analysis, we provide a visualization of a mini-batch from the MNIST dataset as in Figure~\ref{fig:clustered_examples}, showing the clustering results obtained by the GCSD and DDC methods. 
    It reveals that the models trained with DDC and GJRD both often struggle to distinguish between pairs $\left\{ {{\rm{1,8}}} \right\}$, $\left\{ {{\rm{2,7}}} \right\}$, $\left\{ {{\rm{2,8}}} \right\}$, $\left\{{\rm{5,8}} \right\}$, and $\left\{ {{\rm{6,8}}} \right\}$, guiding them to mimic human observers, who often face similar challenges when distinguishing specific pairs of handwritten digits.
    
    \subsubsection{Experiments On Time Series}\label{sec:exp-dynamic} 
    \textbf{Datasets.}
    We conducted a comparative evaluation using three benchmark time-series datasets from the UCI Machine Learning Repository~\cite{asuncion2007uci} and a challenging video dataset derived from a portion of the Dynamic Texture dataset~\cite{peteri2010dyntex}. These four datasets, namely Arabic Digits (AD), Basic Motions (BM), ECG200 (ECG), and Animal Running (AR).
    
    \textbf{Implementation Details.}
    Different from the network on image data, the encoder for the current experiment is implemented using a recurrent network followed by a fully connected layer for the AD, BM, and ECG datasets. Specifically, we employ a two-layer, 32-unit bidirectional Gated Recurrent Unit (GRU)~\cite{cho2014learning} to embed each time series into a vector. 
    To handle video data from the AR dataset, a two-layer 3D CNN with 200 nodes in each layer and convolution filter sizes of $3\times3\times3$ is utilized before the GRU constructs the encoder.
    
    \textbf{Results.}
    Clustering results, as detailed in Table~\ref{tab:DynamicalResults}, demonstrate that the proposed GCSD method outperforms other methods on nearly all the datasets.
    \begin{table}[tbp] %\small
        \centering
        \caption{Clustering Results of Time-series Datasets}
        \label{tab:DynamicalResults}
        \resizebox{\linewidth}{!}{
        \begin{tabular}{@{}lcccccccccc@{}}
            \toprule             
            \multirow{2}{*}{Methods}   & \multicolumn{2}{c}{AD}     & \multicolumn{2}{c}{BM}     & \multicolumn{2}{c}{ECG} & \multicolumn{2}{c}{AR}    \\ \cmidrule(l){2-9}
            & ACC      & NMI           & ACC      & NMI           & ACC      & NMI    & ACC    &NMI    \\ \cmidrule(r){1-9}
            Kmeans~(\cite{hartigan1979algorithm}, 1979)                   & 0.28           & 0.32           & 0.48           & 0.31          & 0.73      & 0.12  & 0.24      & 0.27      \\
            SC~(\cite{shi2000normalized}, 2000)                      & -           & 0.61          & -           & 0.76          & -           & 0.23      & -      & -    \\
            DEC~(\cite{xie2016dec}, 2016)                      & -           & 0.67          & -           & 0.38          & -           & 0.16      & -      & -    \\
            DDC~(\cite{kampffmeyer2019deep}, 2019)                      & 0.80          & \textbf{0.77} & 0.91          & 0.79          & 0.74          & 0.26       & 0.77      & 0.80  \\ \midrule
            GCSD                    & \textbf{0.81}  & 0.75   & \textbf{0.93} & \textbf{0.85} & \textbf{0.79} & \textbf{0.39}  & \textbf{0.80}      & \textbf{0.83} \\ \bottomrule
        \end{tabular}
        }
        \begin{tablenotes}\footnotesize
            \item {
                % The best performances are highlighted in bold. 
                Short underlines ``-" in the table indicate that no results for such dataset or metric are presented in the referenced paper.}
        \end{tablenotes}
    \end{table}

 \section{Application for Domain Adaptation}
    \subsection{Advancing Multi-Source Domain Adaptation} 
    Assume we have a collection of $s$ source domains and one target domain
    $
    \left\{{\mathcal{D}}_1^S,{\mathcal{D}}_2^S,\cdots, {\mathcal{D}}_s^S, {\cal{D}}^T\right\}, 
    $
    each of which follows a specific distribution ${\mathcal{P}}_t$ defined on the space ${\mathcal{X}}\times{\mathcal{Y}}$.
    Suppose we can get access to labeled features for source domain ${\mathcal{D}}_t^S$ with a sample set of $n_t$ $i.i.d.$ samples $(X_t^S, Y_t^S)=\{\mathbf{x}_{t, i}^S, \mathbf{y}_{t, i}^S\}_{i=1}^{n_t}$, but unlabeled features for the target domain with a sample set $X^T=\{\mathbf{x}_{i}^T\}_{i=1}^{n_T}$. 
    The goal of multi-source domain adaptation (MSDA) is to learn a model $f:\mathcal{X} \rightarrow \mathcal{Y}$ with the given data that achieves optimal classification performance on the target domain. 
    
    Pioneers in the field of MSDA have made significant advancements by harnessing the power of deep learning. 
    Their methodologies entail minimizing a loss function that quantifies the discrepancy between the target domain and each source domain~\cite{xu2018deep, zhao2018multiple}. 
    This process facilitates the learning of domain-invariant representations that are universally applicable across all domains. 
    Zhu et al. developed the Multiple Feature Spaces Adaption Network (MFSAN)~\cite{zhu2019aligning} with domain-specific feature extractors and classifiers besides the shared feature extractor for all domains. 
    Peng et al. \cite{peng2019moment} propose the M$^3$SDA to address the MSDA task by dynamically matching not only the moments of feature distributions between each pair of the source and target domains but also the pairs of source domains.
    They tried to train the feature extractor by minimizing the average-pairwise MMD or negative log-likelihood between the feature distributions of different domains and achieved superior performance to existing methods.   
    
    Building upon the concept of M$^3$SDA~\cite{peng2019moment}, the training objective function for a deep learning model in the context of MSDA, incorporating moment matching or aligning of marginal distribution as an objective, can be expressed as follows:
    \begin{equation}\label{eq:objective_DA}
    \begin{array}{c}
    J = \mathop {\min }\limits_{G,C} \;\sum\nolimits_{i = 1}^s {{\rm{CE}}\left( {C\left( {G\left( {X_i^S} \right)} \right),Y_i^S} \right)} \\ 
    + \lambda \sum\nolimits_{i = 1}^s {{\rm{Dis}}\left( {G\left( {X_i^S} \right),G\left( {{X^T}} \right)} \right)} 
    + \gamma \sum\nolimits_{i < j}^S {{\rm{Dis}}\left( {G\left( {X_i^S} \right),G\left( {X_j^S} \right)} \right)} .
    \end{array}
    \end{equation}
    Here, ``$\rm{CE}$'' represents the cross entropy to obtain good classification performance for each source domain, ``$\rm{Dis}$'' is some kind of distance or divergence measure to implement distribution aligning or their moment matching between different domain pairs. 
    $G$ and $C$ denote the feature extractor and classifier, respectively. 
    
    The weights $\lambda$ and $\gamma$ are used to balance the corresponding regularization terms, and different settings for them can help us achieve different goals.
    For instance, if we choose $\gamma=0$, the objective function will prioritize aligning the source domains with the target domain, while disregarding the alignment among the source domains themselves.
    Now let's show how to implement the above objective with our proposed GCSD measure.
    
    If taking $\lambda=\gamma$, then we can calculate the last two parts in Eq.~\eqref{eq:objective_DA} simultaneously with
    \begin{equation}\label{eq:gcsd_msda} 
    \begin{array}{c}
    \lambda (\sum\nolimits_{i = 1}^s {{\rm{Dis}}\left( {G\left( {X_i^S} \right),G\left( {{X^T}} \right)} \right)}  + \sum\nolimits_{i < j}^s {{\rm{Dis}}\left( {G\left( {X_i^S} \right),G\left( {X_j^S} \right)} \right)} ) = :\\
    \lambda {\hat{D}_{{\rm{GCS}}}}\left( {G\left( {X_1^S} \right),\cdots,G\left( {X_s^{S}} \right), G\left( {X^T} \right)} \right),
    \end{array}
    \end{equation}
    where $\hat{D}_{\rm{GCS}}$ can be estimated with Eq.~\eqref{eq:estimator_GCSD}.	
    
    Thus, the objective presented in Eq.~\eqref{eq:objective_DA} can be rewritten as 
    \begin{equation}\label{eq:generalized_div_DA}
    J = \mathop {\min }\limits_{G,C} \;\sum\nolimits_{i = 1}^s {{\rm{CE}}\left( {C\left( {G\left( {X_i^S} \right)} \right),Y_i^S} \right)} 
    + \lambda {\hat{D}_{\rm{GCS}}} (G(X_1^S), \cdots, G(X_{s}^S), G\left( {X^T} \right)).
    \end{equation}
    And we refer to this approach as M$^3$SDA-GCSD.
    \subsection{Experiment Setup}
    We perform an extensive evaluation of the M$^3$SDA-GCSD on the following tasks: digit classification on the Digits-five \cite{peng2019moment} dataset, and image recognition on the Office-31 \cite{saenko2010adapting}, and Office-Home \cite{venkateswara2017deep}.%, OfficeCaltech \cite{zhang2020impact} datasets.\\
    \subsubsection{Datasets} 
    \textbf{Digits-five.} The digits-five dataset is a benchmark for MSDA which includes five distinct datasets: MNIST (MT), SVHN (SV), Synthetic Digits (SD), USPS (US), and MNIST-M (MM), each containing digits ranging from 0 to 9.
    
    \textbf{Office-31.} The Office-31 is a classical domain adaptation benchmark with 31 categories and 4652 images. 
    It contains three domains: Amazon (A), Webcam (W), and DSLR (D), and the data are collected from the office environment.
    
    \textbf{Office-Home.}
    The Office-Home dataset is composed of approximately 15,500 images distributed among 65 categories across four domains: Art (Ar), Clipart (Cl), Product (Pr), and Real World (Rw). Ar contains 2,427 paintings or artistic images from webcam sites, Cl includes 4,365 images from clipart websites, Pr consists of 4,439 webcrawled images, and Rw comprises 4,357 real-world pictures.
    \subsubsection{Baselines and Implementation Details}
    \textbf{Baselines.} 
    There is a substantial amount of MSDA work on real-world visual recognition benchmarks.
    In our experiment, we introduce the MDAN~\cite{zhao2018adversarial}, M$^3$SDA~\cite{peng2019moment}, MFSAN~\cite{zhu2019aligning}, LtC-MSDA~\cite{wang2020learning}, and a recent deep MSDA method CASR~\cite{wang2023class} as baselines. 
    Among them, the MDAN, M$^3$SDA, and MFSAN are marginal distribution matching or aligning methods. 
    
\begin{figure}[tbp]
        \centering
        \includegraphics[width=1.0\linewidth]{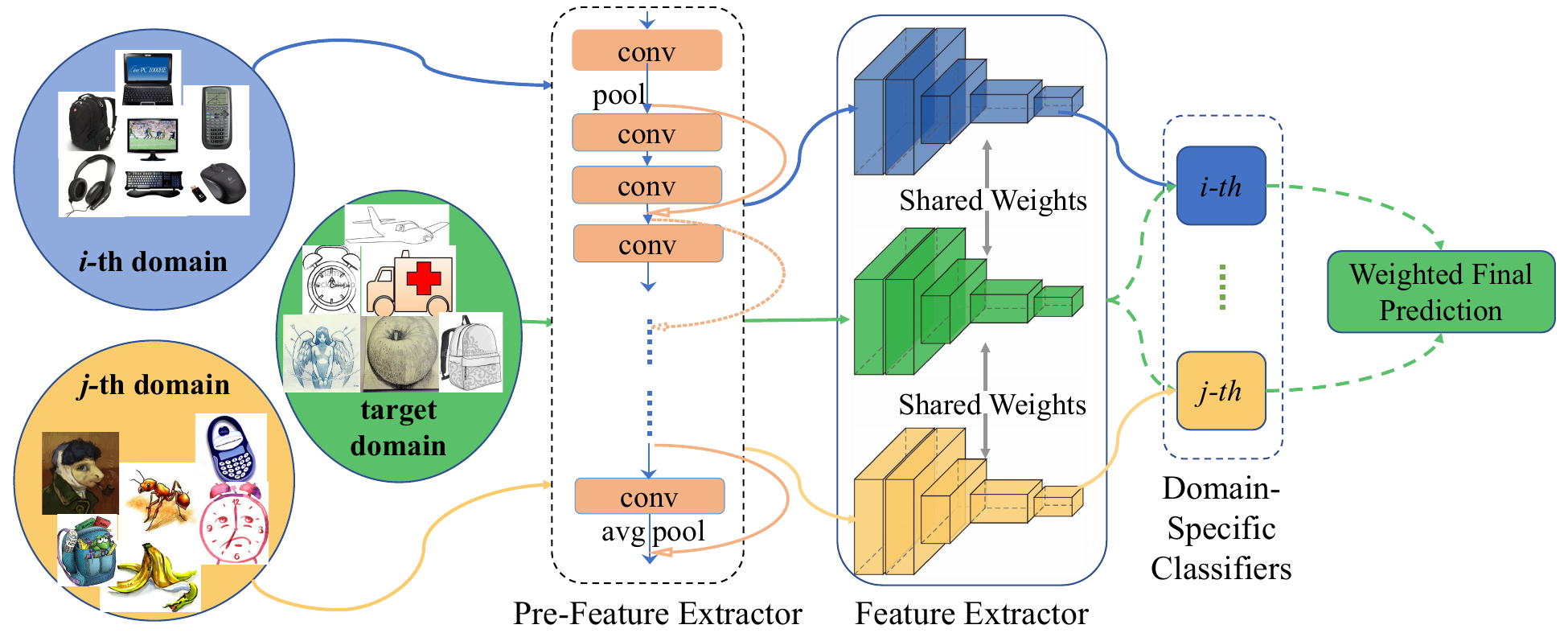}
        \caption{Multi-source domain adaptation framework. 
        The Pre-Feature Extractor is implemented with ResNet50 or ResNet101.
        The Feature Extractor is implemented with CNNs and the Classifiers are Fully Connection Layers. }
        \label{fig:framework-msda}
    \end{figure}

    \textbf{Implementation Details.}
    We followed the setup and training procedure of the M$^3$SDA method for the network architecture and trained them with the proposed objective presented in Eq.~\eqref{eq:generalized_div_DA}.
    The model framework is illustrated in Figure~\ref{fig:framework-msda}.
    
    Specifically, for the Office-31 and the Office-Home dataset, we utilize ResNet-50 and ResNet-101~\cite{he2016deep}, respectively, as a feature extractor without any trainable parameters. 
    Features of the image data are extracted through the ResNet and then fed into a shared extractor $G$,
    which consists of a three-layer conventional network with hidden nodes $512-256-128$, convolution filter size 3, stride 2, and padding 1. 
    Each conventional layer is followed by batch-normalization and ReLU activation and a $3\times3$ max pooling layer. 
    
    For the Digits-Five dataset, we use no pre-feature extractor but only a three-layer conventional ($conv$) followed by two fully connected linear ($fc$) for the shared feature extractor $G$, whose architecture is consistent with the M$^3$SDA.
    
    All the experiments are conducted with a learning rate $2.0e^{-4}$, maximum epochs 100, batch size 128 for Digits-Five, and 64 for other datasets. 
    During training, the regularization weight $\lambda$ for $D_*$ is set to 0.5 for all the experiments.
    
    \subsection{Results and Comparison}
    The classification results on three benchmark datasets are shown in Table~\ref{tab:Digits-five}, \ref{tab:Office-31}, and \ref{tab:Office-Home}. 
    As can be seen, M$^3$SDA-GCSD achieves superior performance in most of the transfer tasks. 
    In terms of average accuracy, it outperforms all the compared baseline methods by significant margins, particularly in classification tasks involving more classes.
    
    It is worth noting that the M$^3$SDA-GCSD improves the average classification accuracy quite a lot from the original M$^3$SDA.
    Specifically, the improvement approaches 7.6\% for the Digits-five dataset, 11.1\% for the Office-31 dataset, and 11.3\% for Office-Home.
    Additionally, it is found that our method significantly reduces the performance deviation across different transfer tasks.
    This implies that the proposed method outperforms competing approaches in achieving better generalization across different domains.
    
    The observed improvements in both average accuracy and its deviation demonstrate that minimizing the generalized divergence measure effectively aligns the feature distributions of the given domains, resulting in significant domain adaptation and generalization capability.
    This finding suggests that there is significant potential to employ generalized divergence measures like the proposed GCSD to align feature distributions from different domains.
    \begin{table}[]\centering
        \caption{Classification accuracy (\%) on Digits-five dataset.}
        \label{tab:Digits-five}
        % \resizebox{\columnwidth}{!}{%
        \begin{tabular}{@{}lcccccc@{}}
            \toprule
            Methods                         & $\to$MM   & $\to$MT   & $\to$US   & $\to$SV   & $\to$SD    & AVG \\ \midrule
            MDAN~(\cite{zhao2018adversarial}, 2018) & 69.5       & 98        & 92.4      & 69.2      & 87.4      & 83.3$\pm$13.3\\
            % \textcolor{red}{MFSAN}        & 69.5       & 98        & 92.4      & 69.2      & 87.4      & 83.3 \\
            M$^3$SDA~(\cite{peng2019moment}, 2019)  & 72.8       & 98.4      & 96.1      & 81.3      & 89.6      & 87.7$\pm$10.6 \\
            LtC-MSDA~(\cite{wang2020learning}, 2020) & 85.6       & 99        & 98.3      & 83.2      & 93        & 91.8$\pm$7.2 \\
            CASR~(\cite{wang2023class}, 2023)       & 90.2  & \textbf{99.7}  & 98.3      & 86.4      & 96.3      & 94.1$\pm$5.7 \\ \midrule
            % MOST~\cite{nguyen2021most}    & 91.5       & 99.6      & 98.4      & 90.9      & 96.4      & 95.4 \\\midrule
            M$^3$SDA-GCSD     & \textbf{95.6}       & 99.0      & \textbf{99.3}      & \textbf{85.0}      & \textbf{97.5}      & \textbf{95.3$\pm$5.9} \\  \bottomrule
        \end{tabular}
        % }
        \begin{tablenotes}\footnotesize
            \item {``$\to$ * '' denotes a transfer task from domains without * to *.
            Our M$^3$SDA-GCSD achieves 95.3\% average accuracy, outperforming other baselines by a large margin.}
        \end{tablenotes}
    \end{table} 
    
    \begin{table}[]
    \centering
    \caption{Classification accuracy (\%) on Office-31 dataset.}
    \label{tab:Office-31}
    \begin{tabular}{@{}lcccc@{}}
        \toprule
        Methods                         & $\to$A &  $\to$W &    $\to$D      & AVG \\ \midrule
        MDAN~(\cite{zhao2018adversarial}, 2018)& 55.2      & 95.4      & 99.2      & 83.3$\pm$24.4\\
        M$^3$SDA~(\cite{peng2019moment}, 2019)  & 55.4      & 96.2      & 99.4      & 83.7$\pm$24.5\\
        MFSAN~(\cite{zhu2019aligning}, 2019)    & 72.5      & 98.5      & 99.5      & 90.2$\pm$20.2\\
        LtC-MSDA~(\cite{wang2020learning}, 2020) & 63.9      & 98.4      & 99.2      & 87.2$\pm$24.0\\
        CASR~(\cite{wang2023class}, 2023)      & 76.2      & 99.8      & \textbf{99.8}      & 91.9$\pm$13.6\\ \midrule
        M$^3$SDA-GCSD      & \textbf{84.6}		& \textbf{100}       & 99.6      & \textbf{94.8$\pm$8.8}  \\ \bottomrule
    \end{tabular}
    \end{table}
    
    % Please add the following required packages to your document preamble:
    % \usepackage{booktabs}
    \begin{table}[]
    \centering
    \caption{Classification accuracy (\%) on Office-Home dataset.}
    \label{tab:Office-Home}
    \begin{tabular}{@{}lccccc@{}}
        \toprule
        Methods                         & $\to$A  & $\to$C &  $\to$P & $\to$R  & AVG    \\ \midrule
        MDAN~(\cite{zhao2018adversarial}, 2018) & 64.9  & 49.7    & 69.2     & 76.3    & 65.0$\pm$11.2\\
        M$^3$SDA~(\cite{peng2019moment}, 2019)  & 64.1  & 62.8    & 76.2     & 78.6    & 70.4$\pm$8.1   \\
        % MADAN      & 66.8             & 54.9   & 78.2   & 81.5     & 70.4    &12.1   \\
        MFSAN~(\cite{zhu2019aligning}, 2019)    & 72.1  & 62      & 80.3     & 81.8    & 74.1$\pm$9.1 \\
        CASR~(\cite{wang2023class}, 2023)       & 72.2  & 61.1    & 82.8     & 82.8    & 74.7$\pm$10.4 \\ \midrule
        M$^3$SDA-GCSD                   & \textbf{78.3}   & \textbf{74.2}	  & \textbf{89.5}	 & \textbf{84.7}    & \textbf{81.7$\pm$6.7 }      \\ \bottomrule
    \end{tabular}
    \end{table}  
    
\section{Conclusion}
To overcome the dilemma of lacking effective and efficient metrics when dealing with data from multiple distributions in diverse deep learning applications such as clustering, multi-source domain adaptation, multi-view learning, and multi-task learning, we propose a novel divergence measure termed the Generalized Cauchy-Schwarz Divergence (GCSD). 
This measure incorporates closed-form empirical estimation, rendering it well-suited and convenient for integration into machine-learning frameworks.
The effectiveness and efficiency are demonstrated by computational complexity analysis and power testing.
Extensive evaluations conducted on two demanding tasks, deep clustering and multi-source domain adaptation, lead to two noteworthy conclusions.
Firstly, both GCSD-based clustering and domain adaptation frameworks surpass state-of-the-art baselines across diverse benchmark datasets, underscoring the remarkable capacity of GCSD in quantifying the divergence among multiple distributions.
Furthermore, GCSD is applicable in various deep-learning scenarios involving the quantification of generalized divergence between multiple distributions, regardless of whether the objective is maximizing divergence for feature discrimination or minimizing it for feature alignment.
    
%\end{linenumbers}
\section*{Acknowledgements}
This work was partially funded by the National Natural Science Foundation of China, grant no. U21A20485 and 62088102, as well as the Research Council of Norway, grant no. 309439, Visual Intelligence Centre.
\bibliographystyle{elsarticle-num}
% \bibliography{ref.bib}	
\bibliography{ref-styled}	

\newpage
\appendix % 切换到附录模式

\newcounter{appprop}
\newtheorem{appProp}[appprop]{Proposition} %create gloabal theorem environment
    
\section{Details for Power Test Experiment}\label{app:powertest}
        We generated datasets from synthetic multiple distributions to evaluate the effectiveness of our proposed divergence measures and three pairwise versions of competitors, namely pairwise Kullback–Leibler Divergence (pKLD), pairwise Maximum Mean Divergence (pMMD), and pairwise Cauchy-Schwarz Divergence (pCSD). 
        The pairwise divergence measure is defined as the average of pairwise divergences between each two of all the $m$ distributions. 
        \begin{equation}
        {D_*}\left( {\left\{ {{P_i}} \right\}_1^m} \right) = \sum\limits_{i, j}^m {{d_*}\left( {{P_i},{P_j}} \right)} ,
        \end{equation}
        where $d_*$ can be referred to KLD, MMD, CSD, and any other existing divergence measures.
        It is important to note that the MMD is a two-sample divergence estimator. 
        In contrast, both the KLD and the CSD are defined on continuous space for distributions. 
        Therefore, we implement both of them with KDE.
        
\begin{figure}
            \centering
            \includegraphics[width=0.5\linewidth]{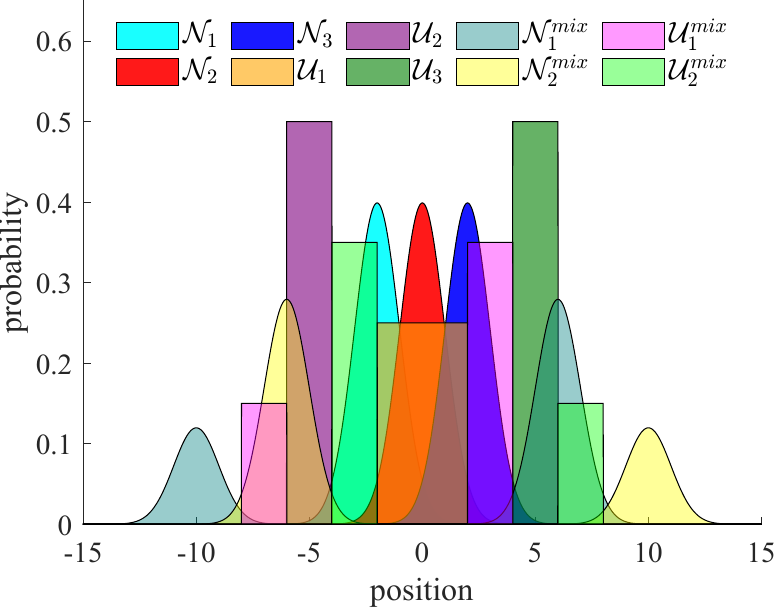}
            \caption{Test data visualization}
            \label{fig:syn_distributions}
        \end{figure}

        The synthetic dataset includes ten distributions with different density centers, the range of which is scaled by a given value $r$.
        The ten distributions consisted of three Gaussians, three Uniforms, two bimodal Gaussians, and two bimodal Uniforms, with the corresponding probability density functions as follows:
        \begin{equation}
        \begin{array}{l}
        \begin{array}{*{20}{l}}
        {{f_1}\left( x \right) = {\cal N}\left( {0,1} \right)}
        &{{f_2}\left( x \right) = {\cal N}\left( {s,1} \right)}
        &{{f_3}\left( x \right) = {\cal N}\left( { - s,1} \right)}\\
        {{f_4}\left( x \right) = {\cal U}\left( { - s,s} \right)}
        &{{f_5}\left( x \right) = {\cal U}\left( { - 3s, - 2s} \right)}
        &{{f_6}\left( x \right) = {\cal U}\left( {2s,3s} \right)}
        \end{array}\\
        \begin{array}{*{20}{l}}
        {{f_7}\left( x \right) = 0.3{\cal N}\left( { - 5s,1} \right) + 0.7{\cal N}\left( {3s,1} \right)}
        &{{f_{8}}\left( x \right) = 0.3{\cal N}\left( { - 3s,1} \right) + 0.7{\cal N}\left( {5s,1} \right)}\\
        {{f_9}\left( x \right) = 0.3{\cal U}\left( { - 4s, - 3s} \right) + 0.7{\cal U}\left( {s,2s} \right)
        }&{{f_{10}}\left( x \right) = 0.3{\cal U}\left( {3s,4s} \right) + 0.7{\cal U}\left( { - 2s, - s} \right)}
        \end{array}
        \end{array}.
        \end{equation}  
        Here, $s = \frac{r}{10}$, and ${\cal N}(\mu, \sigma)$ represents the probability density function of a normal distribution with mean $\mu$ and standard deviation $\sigma$. 
        Similarly, $U(l,u)$ denotes the Uniform distribution with $l$ and $u$ as the lower and upper bounds, respectively.        
        Visualization of the synthetic data with scatter range $r=20$ (The distance from the leftmost density center to the rightmost density center) is shown in Figure~\ref{fig:syn_distributions}.  
    
\begin{figure}[htp]
            \centering
            \subfloat[Normalized divergence]{\centering\includegraphics[width=0.48\linewidth]{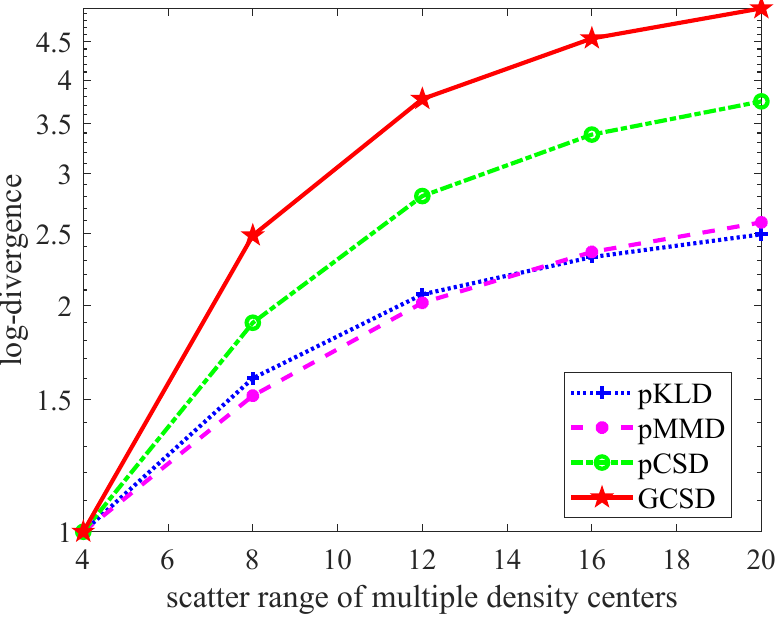} \label{fig:powertest_divs}}
            \hfill        
            \subfloat[Time consumption]{\centering\includegraphics[width=0.48\linewidth]{time_consumption}
                \label{fig:powertest_time_consumption}}
            \caption{Divergences measured and time consumption.}            
            \label{fig:power_test_semilog}
        \end{figure}

        Intuitively, the parameter $r$ controls the scattering range of the multi-distribution, which means that the sample sets from such synthetic distribution combinations with a larger $r$ value are more dispersed, thus resulting in a larger divergence.
        To test the effectiveness of our proposed divergence measures, we generated five sample sets using the multi-distribution defined above with scaling ranges of $r\in\left\{4,8,12,16,20\right\}$. 
        Each distribution contributed 2000 samples, making the total size of the test dataset to be size 20000. 
        We then conducted divergence tests on these five sample sets using the proposed GCSD and the three competing pairwise divergences mentioned earlier.
        % Results are shown in Figure~\ref{fig:power_test_semilog}. 
        Figure~\ref{fig:powertest_divs} shows the divergence measured by various metrics and Figure~\ref{fig:powertest_time_consumption} depicts the time consumption by them.

\section{Deduction of Proposition \ref{prop1}}\label{app:proof} 
    Continuing with the deduction presented in Eq.~\eqref{eq:cross_entropy_gcsd}$\sim$\eqref{eq:power_entropy_gcsd}, we derive
    \begin{equation}\label{eq:cross_entropy_gscd_cluster}
    \begin{aligned}         
        {\hat V}_1 
        &= {\frac{1}{m}\sum\nolimits_{t = 1}^m {\frac{1}{{{n_t}}}\sum\nolimits_{j = 1}^{{n_t}} {\left( {\prod\nolimits_{k \ne t}^m {\frac{1}{{{n_k}}}\sum\nolimits_{i = 1}^{{n_k}} {{\kappa _\sigma }\left( {{\rm{x}}_j^t - {\rm{x}}_i^k} \right)} } } \right)} } } \\
        &=  {\frac{1}{m}\sum\nolimits_{t = 1}^m {\frac{1}{n}\sum\nolimits_{j = 1}^n {\left( {\prod\nolimits_{k \ne t}^m {{a_{j,t}}\frac{1}{n}\sum\nolimits_{i = 1}^n {{a_{i,k}}{\kappa _\sigma }\left( {{{\rm{x}}_j} - {{\rm{x}}_i}} \right)} } } \right)} } }\\
        & =  {\frac{1}{{m{n^m}}}\sum\nolimits_{t = 1}^m {\sum\nolimits_{j = 1}^n {\left( {a_{j,t}^{m - 1}{\rm{prod}}\left( {{K_j}A} \right)/{K_j}{\alpha _t}} \right)} } }\\       
        & =  {\frac{1}{{m{n^m}}}{\rm{sum}}\left( {\frac{{{{\left( {{A^{m - 1}}} \right)}^T}}}{{KA}}{\rm{prod}}\left( {KA} \right)} \right)},
    \end{aligned}
    \end{equation}        
    and 
    \begin{equation}\label{eq:power_entropy_gscd_cluster}
    \begin{aligned}
        \frac{1}{m}\sum\nolimits_{t = 1}^m {\log \left( {\hat V}_2 \right) }
        &= \frac{1}{m}\sum\nolimits_{t = 1}^m {\log \left( {\frac{1}{{{n_t}}}\sum\nolimits_{j = 1}^{{n_t}} {p_t^{m - 1}\left( {{{\rm{x}}_j}} \right)} } \right)} \\
        &= \frac{1}{m}\sum\nolimits_{t = 1}^m {\log \left( {\frac{1}{{{n_t}}}\sum\nolimits_{j = 1}^{{n_t}} {{{\left( {\frac{1}{{{n_t}}}\sum\nolimits_{i = 1}^{{n_t}} {{\kappa _\sigma }\left( {{\rm{x}}_j^t - {\rm{x}}_i^t} \right)} } \right)}^{m - 1}}} } \right)} \\
        &= \frac{1}{m}\sum\nolimits_{t = 1}^m {\log \left( {\frac{1}{{{n^m}}}\sum\nolimits_{j = 1}^n {{{\left( {\sum\nolimits_{i = 1}^n {{a_{i,t}}{a_{j,t}}{\kappa _\sigma }\left( {{{\rm{x}}_j} - {{\rm{x}}_i}} \right)} } \right)}^{m - 1}}} } \right)} \\
        &= \frac{1}{m}\sum\nolimits_{t = 1}^m {\log \left( {\frac{1}{{{n^m}}}\sum\nolimits_{j = 1}^n {{{\left( {{a_{j,t}}{K_j}{\alpha _t}} \right)}^{m - 1}}} } \right)} \\
        &= \frac{1}{m}{\rm{tr}}\left( {\log \left( {\frac{1}{{{n^m}}}{{\left( {{A^T}} \right)}^{m - 1}}{{\left( {KA} \right)}^{m - 1}}} \right)} \right).
    \end{aligned}
    \end{equation}
    Here and after, $K$ represents the Gram matrix obtained by evaluating the positive definite kernel ${\kappa _\sigma }$ on all sample pairs, such that $K_{i,j} = {\kappa _\sigma }({\bf x}_i-{\bf x}_j)$. 
    The notation ${\rm{sum}}(\cdot)$ signifies the summation of all elements within a matrix. 
    The function ${\rm{prod}}(A)$ calculates the product of row elements in matrix $A$, yielding a column vector, i.e., ${\rm{prod}}(A)=\left[ {\prod\nolimits_{j = 1}^m {{A_{1,j}}} , \cdots ,\prod\nolimits_{j = 1}^m {{A_{n,j}}} }\right]^T$. 
    % The function ${\rm{rep}}(\cdot)$ generates $m$ copies of a given column vector and concatenates them to form a new matrix. 
    ${\rm{tr}}(\cdot)$ denotes the trace of the provided matrix. 
    Additionally, the symbols $A^{n}$, and $\frac{A}{B}$ represent element-wise exponentiation, and division for matrices, respectively.
    
    Take Eq.~\eqref{eq:cross_entropy_gscd_cluster} $\sim$ \eqref{eq:power_entropy_gscd_cluster} into Eq.~\eqref{gcsd_def}, we obtain the Proposition \ref{appprop1} as below.    
    \begin{appProp}\label{appprop1}
        Given data samples $\left\{\mathbf{x}_i\right\}_{i=1}^n \in \mathbb{R}^d$ as well as their cluster-assignment matrix $A\in \mathbb{R}^{n \times m}$, the generalized Cauchy-Schwarz divergence of the dataset based on the assignment matrix can be computed as:
        \begin{equation}\label{eq:cluster_GCSD_app}
          \begin{aligned}
            {\hat D}_{\rm{GCS}}^{A}({\cal P}_{1:m}) =  
            &- \log \left( {\frac{1}{m}{\rm{sum}}\left( {\frac{{{{\left( {{A^{m - 1}}} \right)}^T}}}{{KA}}{\rm{prod}}\left( {KA} \right)} \right)} \right) \\
            &+ \frac{1}{m}{\rm{tr}}\left( {\log \left( {{{\left( {{A^T}} \right)}^{m - 1}}{{\left( {KA} \right)}^{m - 1}}} \right)} \right).              
          \end{aligned}
        \end{equation}
    \end{appProp}      
       
\end{sloppypar}
\end{document}